\documentclass[pdflatex,sn-aps]{sn-jnl}% American Physical Society (APS) Reference Style
\usepackage{graphicx}%
\usepackage{multirow}%
\usepackage{amsmath,amssymb,amsfonts}%
\usepackage{amsthm}%
\usepackage{mathrsfs}%
\usepackage[title]{appendix}%
\usepackage{xcolor}%
\usepackage{textcomp}%
\usepackage{manyfoot}%
\usepackage{booktabs}%
\usepackage{algorithm}%
\usepackage{algorithmicx}%
\usepackage{algpseudocode}%
\usepackage{listings}%
\usepackage{rotating}
\usepackage{longtable}
\usepackage{pdflscape}
\usepackage{changepage}
\usepackage{afterpage}
\usepackage{array}
\usepackage{ragged2e}
\usepackage{lscape}
\usepackage{fancyhdr}
\usepackage{array}
\usepackage{caption}
\usepackage{subfig}
\usepackage{tabularx}
\usepackage{adjustbox}
\usepackage{natbib}
\usepackage{placeins}
\usepackage{geometry}
\usepackage{float}
\fancypagestyle{mylandscape}{
  \fancyhf{}
 
  \fancyfoot[C]{\thepage}
}
\begin{document}

\title[Short title]{From Automation to Autonomy in Smart Manufacturing: A Bayesian Optimization Framework for Modeling Multi-Objective Experimentation and Sequential Decision Making}


%% Author definitions
\author[1]{\fnm{Avijit Saha} \sur{Asru}}\email{avijit.saha@buft.edu.bd}
\author[2]{\fnm{Hamed} \sur{Khosravi}}\email{hamedkhosravi181@gmail.com}
\author*[2]{\fnm{Imtiaz} \sur{Ahmed}}\email{imtiaz.ahmed@mail.wvu.edu} % Corresponding author
\author[1]{\fnm{Abdullahil} \sur{Azeem}}\email{azeem@ipe.buet.ac.bd}

%% Affiliations
\affil[1]{\orgdiv{Department of Industrial and Production Engineering}, \orgname{Bangladesh University of Engineering and Technology}, \orgaddress{\city{Dhaka}, \country{Bangladesh}}}
\affil[2]{\orgdiv{Department of Industrial and Management Systems Engineering}, \orgname{West Virginia University}, \orgaddress{\city{Morgantown}, \state{West Virginia}, \country{USA}}}


\abstract{Discovering novel materials with desired properties is essential for driving innovation across industries. Industry 4.0 and smart manufacturing have promised transformative advances in this area through real-time data integration and automated production planning and control. However, the reliance on automation alone has often fallen short, lacking the flexibility and innovation needed for this complex process. Automation, constrained by predefined processes, is unable to adapt to real-time changes. To fully unlock the potential of smart manufacturing, we must evolve from automation to autonomous systems that go beyond rigid programming and can dynamically adjust and optimize the search for solutions in real time. Current discovery approaches are often slow, requiring numerous trials to find optimal combinations, and costly, particularly when optimizing multiple properties simultaneously. To address this challenge, this paper proposes a Bayesian multi-objective sequential decision-making (BMSDM) framework that can intelligently select experiments as manufacturing progresses, guiding us toward the discovery of optimal design faster and more efficiently. The framework leverages sequential learning through a Bayesian Optimization (BO) framework, which iteratively refines a statistical model representing the underlying manufacturing process. This statistical model, a Gaussian Process (GP), acts as a surrogate, allowing for efficient exploration and optimization without requiring numerous real-world experiments. This approach has the potential to significantly reduce the time and cost of data collection required by traditional experimental designs. To prove our hypothesis, the proposed BMSDM is compared with traditional Design of Experiments (DoE) methods and two state-of-the-art multi-objective optimization methods. Using a real manufacturing dataset, we evaluate and compare the performance of these approaches across five key evaluation metrics. Our results demonstrate that BMSDM comprehensively outperforms the competing methods in multi-objective decision-making (MODM) scenarios. Our proposed approach represents a significant leap forward in creating a futuristic intelligent autonomous platform capable of novel material discovery, moving from rigid automation to adaptive, autonomous systems.}

\maketitle
\captionsetup[table]{labelsep=space} % Remove colon
\captionsetup[figure]{labelsep=space} % Remove colon
\section{Introduction}
\label{Intro}

%% Use \subsection commands to start a subsection.
Since the onset of industrialization, successive technological breakthroughs have precipitated major shifts, historically framed as ``Industrial Revolutions" (IR) \citep{Lasi2014}. The latest of these revolutions, known as ``Industry 4.0," merges the Internet of Things (IoT) with cyber-physical systems, creating an integrated platform capable of receiving and processing information to enhance manufacturing operations. This era leverages cloud computing, IoT, big data analytics, and artificial intelligence (AI) to foster the development of intelligent systems for real-time monitoring and control of manufacturing processes \citep{bahrin2016industry, vaidya2018industry, zheng2018smart}. Coined as ``Smart Manufacturing" (SM), this approach is fundamentally data-driven, utilizing IoT devices and various sensors to collect and analyze data throughout a product's lifecycle, thereby promoting self-learning and adaptive capabilities in manufacturing processes \citep{Tao2018, Wang2018, ghahramani2020ai}.

Building on the technological foundation set by Industry 4.0, machine learning (ML) emerges as a pivotal element in driving the intelligence of manufacturing processes \citep{liu2017materials}. ML is essential in Industry 4.0, facilitating automation, enhancing efficiency, and promoting intelligent decision-making through the utilization of industrial big data and AI-driven analytics. With the recent integration of large language models (LLMs) into ML frameworks, their influence has expanded across diverse industrial domains. The Industrial Large Knowledge Model (ILKM), when combined with ML, strengthens Industry 4.0 by seamlessly incorporating AI-driven analytics with industrial big data, optimizing decision-making, process automation, and predictive maintenance. By harnessing deep learning, knowledge graphs, and real-time data processing, ILKM facilitates the development of intelligent, scalable, and autonomous industrial systems \citep{lee2024unified}. It has been transformative, driving the data-driven evolution from traditional manufacturing techniques to cutting-edge innovations such as additive manufacturing (AM) and novel material discovery \citep{Guillen2020, liu2017materials}. Even beyond manufacturing processes, advancements in ML for industrial system diagnostics and prognostics \citep{su2023machine} have highlighted the potential of data-driven learning in refining predictive models, enhancing fault detection, and improving system reliability, further reinforcing the need for robust and adaptive ML frameworks in SM. However, despite the rapid advancements, traditional ML models often struggle to meet the dynamic demands of data-driven SM. They show a vulnerability to slight variations in manufacturing processes \citep{Ramezankhani2021} and a high dependence on substantial volumes of labeled data \citep{botcha2021efficient}.

In the dynamic and complex world of manufacturing, the interplay of process parameters significantly influences the efficacy and quality of outputs \citep{rao2018multi}. As industries strive for peak performance, the optimization of these parameters becomes crucial, leveraging cutting-edge routines to discover optimal settings \citep{ruane2023using}. The fusion of data analytics with optimization processes has notably advanced the intelligent capabilities of various industrial operations \citep{tang2021data}. Whether dealing with single-objective goals to reduce costs and maximize efficiency \citep{fountas2020single} or navigating the intricate challenges of multi-objective optimization (MOO) where competing goals must be balanced \citep{daulton2020differentiable}, the need for innovative approaches is more pressing than ever. In manufacturing contexts, single-objective optimization often focuses on minimizing production costs or maximizing output efficiency \citep{fountas2020single}. However, materials used in various applications usually need to meet diverse property requirements, complicating the optimization process \citep{talapatra2018autonomous}. In scientific and engineering experimental design, the challenge often lies in simultaneously optimizing multiple competing objectives, which are frequently interlinked through complex, opaque functions \citep{konakovic2020diversity}. Such scenarios are prevalent in fields like chemical design, battery optimization, AM, clinical drug trials, process parameter selection, and novel material design \citep{konakovic2020diversity}.

Particularly in AM and novel material design, the variability in process parameters and extensive search space can lead to significant discrepancies in component quality and high turnaround time, respectively, prompting extensive research into optimal parameter settings through meticulous experimental designs \citep{kumar2023machine}. Given the high costs and resource-intensive nature of physical experiments, traditional DoE methods are often employed to maximize the exploration of design spaces efficiently \citep{greenhill2020bayesian, montgomery2017design}. These approaches can be prohibitively expensive and rigid, lacking the flexibility to adapt to new data or complex, dynamic systems. Furthermore, metaheuristic approaches like the Multi-Objective Genetic Algorithm (MOGA) \citep{murata1995moga}, Non-dominated Sorting Genetic Algorithm (NSGA-II) \citep{deb2002fast}, Multi-Objective Artificial Bee Colony (MOABC) \citep{akbari2012multi}, and Multi-Objective Particle Swarm Optimization (MOPSO) \citep{patil2014multi} have also been explored for tackling data-driven optimization challenges, such as novel material design, with fewer evaluations. While they offer a powerful and robust optimization framework, they typically require large datasets and struggle to effectively balance conflicting objectives in the absence of extensive training data \citep{gao2000study, patil2014multi}.

Addressing these limitations, recent advancements in active learning and sequential design frameworks provide a strategic edge, particularly in domains characterized by smaller datasets such as new material development and clinical trials \citep{botcha2021efficient, zerka2021privacy}. Sequential learning is a process where a computational agent continually seeks new data to update a model, balancing exploitation of existing knowledge with exploration of previously unknown areas \citep{Palizhati2022}. This approach is crucial in real-world scenarios such as robotics, automation, and gaming, where making optimal decisions with limited resources is essential \citep{li2024multi}. Among sequential learning methods, BO \citep{frazier2018tutorial} particularly stands out by addressing the core disadvantages of both traditional DoE and heuristic-based methods \citep{daulton2022multi, Hanaoka2022}. It offers a cost-effective alternative by requiring fewer experiments to achieve comparable or superior outcomes, making it economically viable for many industrial applications \citep{ahmed2021,raihan2024,khosravi2024data}. Moreover, BO’s model-updating capability provides a foundation for continuous improvement and autonomous decision-making in manufacturing processes, marking a significant step towards fully autonomous manufacturing platforms. However, the Gaussian Process (GP) model, which is used in BO to approximate the objective function, struggles with discontinuities because it assumes smooth and continuous objective functions \citep{picheny2019ordinal}.  Standard GP models presume stationarity, indicating that the function's smoothness is uniform throughout the input space. Nonetheless, practical optimization problems frequently exhibit areas of differing smoothness \citep{sauer2023non}.

%This methodology not only streamlines design processes but also enhances adaptability, allowing manufacturing systems to respond to changes in real-time and continuously improve without extensive human intervention.

\begin{figure}[h]
    \centering
    \includegraphics[height=4.5in,keepaspectratio]{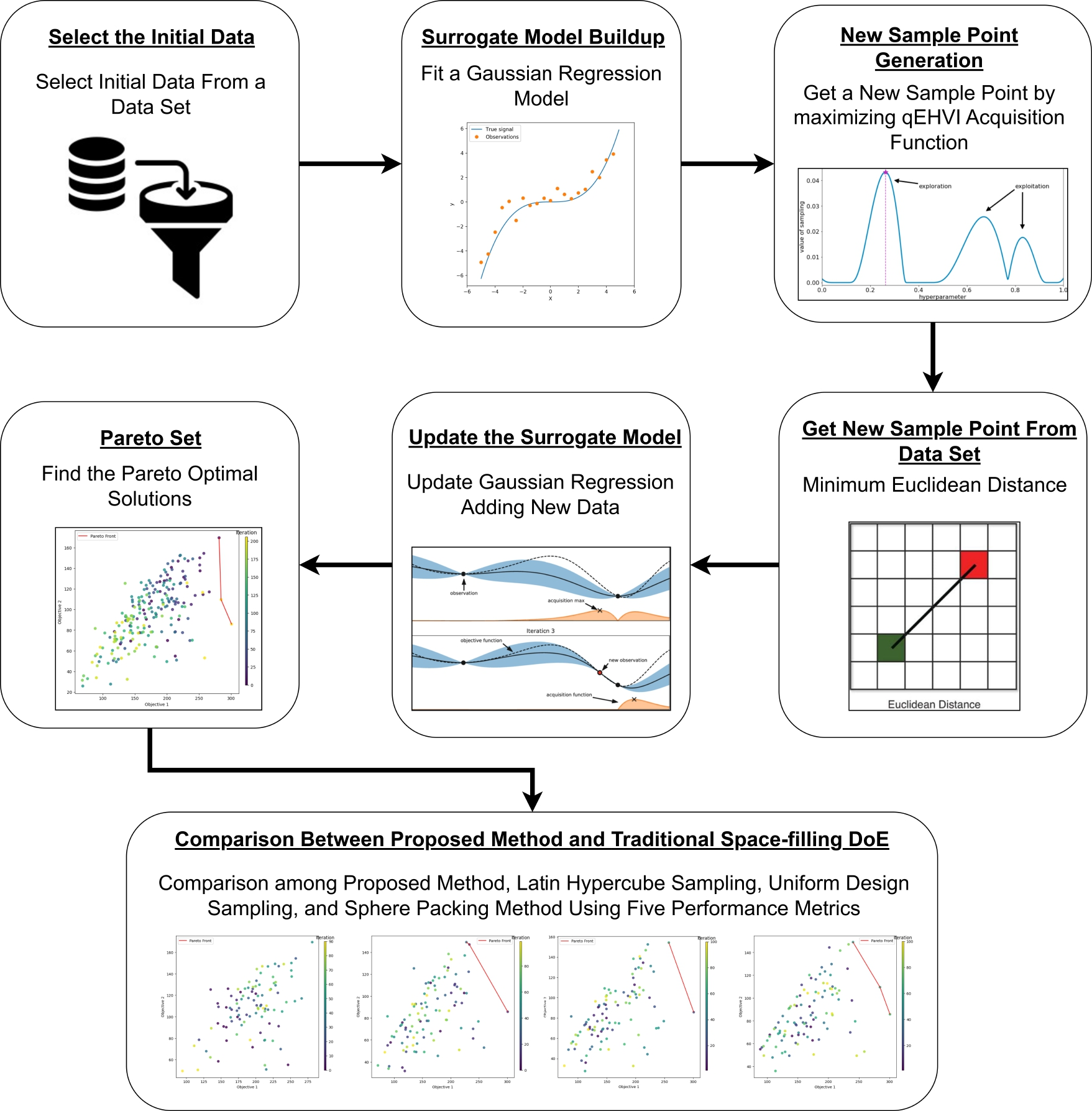}
    \caption{The Overview of the Study}
    \label{study_overview}
\end{figure}

This study dives into the sophisticated mechanisms of  BO-based sequential designs, demonstrating how they surpass traditional models by seamlessly integrating new data and adapting strategies in real time. Rather than relying on preset rules and requiring human intervention to adapt to changes, the data-driven framework presented in this study enables systems to learn from real-time data, make autonomous decisions, and continuously optimize their operations. With this self-evolving capability, the system transitions from rigid automation to a flexible, intelligent, and autonomous system, dramatically enhancing efficiency and adaptability in manufacturing environments.  The contributions of this work are summarized below:
\begin{itemize}
    \item We introduce a novel and efficient BO  framework, BMSDM, for handling MODM scenarios. It is designed to intelligently and iteratively determine the optimal combinations of process parameters. 
    %This approach significantly accelerates the attainment of the optimal design by making progressively informed decisions throughout the optimization process. 
    \item Within our BO framework, we utilize a batch-wise hypervolume improvement-based acquisition function to determine each experiment's location in successive iterations. This approach enables us to continually update and refine a GP based surrogate model representing the manufacturing process. 
    \item We compare conventional DoE approaches such as Latin Hypercube Sampling (LHS) \citep{husslage2011space, joseph2016space}, Uniform Design Sampling (UDS) \citep{harada2007uniform}, and Sphere Packing Method (SPM) \citep{hifi2019local} with the proposed framework using five different performance metrics: Generational Distance (GD), Inverted Generational Distance (IGD), Hypervolume (HV), Proportional Hypervolume (PHV) and percentage of Data Usage (D).
    \item We utilize a real-life manufacturing dataset \citep{talapatra2018autonomous} to simulate a dynamic learning environment. We explore two scenarios: one where both objective functions are to be maximized, and another where one objective function is to be maximized while the other is to be minimized. The overview of this study is summarized in Fig.~\ref{study_overview}.
    \item We further evaluate the performance of our proposed approach against two state-of-the-art MOO methods across various performance metrics.
\end{itemize}

The rest of the paper is organized as follows: Section \ref{Literature} presents an overview of the literature to ascertain the versatility of MOO problems and how classical, metaheuristic algorithms, sequential methods, and BO have evolved. Section \ref{Methodology} outlines the preliminary work, definitions, procedures, and performance measures of our proposed technique. The manufacturing dataset, simulation, comparison study, and computational results are highlighted in Section \ref{Result}. The study's limitations, recommendations, and closing thoughts are provided in Section \ref{Conclusion}, which also serves as a summary of the work.

%\appendix
\section{Literature Review}
\label{Literature}
Current MOO techniques are categorized into three distinct types, excluding those that address uncertainty \citep{wang2004efficient}. The first type involves transforming MOO problems into single-objective problems by employing weights, preferences or aims \citep{tappeta2000interactive}. The second type focuses on determining Pareto set points without considering preferences \citep{marler2010weighted}. The third type models goal functions using surrogate models or by directly approximating the Pareto optimal frontier, rather than engaging in computationally intensive studies \citep{lotov2008visualizing, wilson2001efficient}. A significant challenge with the first approach is the difficulty in selecting appropriate weighting elements for practical application in real-world problems. Evolutionary algorithms are notably the most effective and widely used method for the second approach \citep{deb2003towards}, providing decision-makers with numerous Pareto set points. However, these techniques are typically computationally expensive due to the extensive number of points that need evaluation \citep{wang2004efficient}.

In this section, the literature review is organized into two main subsections. The first subsection explores space-filling techniques used in the DoE methodology and traces the evolution of MOO methods from metaheuristic approaches to BO processes. The second subsection discusses the recent advancements in BO, specifically its application in solving costly multi-objective black-box optimization challenges.

\subsection{Comprehensive Review of Relevant Literature of DoE Techniques \& MOO Methods}
DoE and space-filling designs are critical initial steps in effectively planning and executing experiments, particularly when addressing MOO problems. The primary function of the DoE is to strategically design experiments so that the data collected are systematic and comprehensive, covering a wide range of possible experimental conditions. This structured approach ensures that all relevant variables are thoroughly investigated across the experimental domain. Space-filling designs, including methods like LHS, UDS, and SPM, play a key role in this process. These designs are employed to distribute experiment points uniformly throughout the design space, minimizing gaps and maximizing coverage \citep{montgomery2017design}.

Once the experimental locations are determined through DoE and space-filling methods, the resultant data provide a robust foundation for solving MOO problems. An outline of developments in DoE and MOO approaches is provided in Table~\ref{tab:optimization_study}.

\begin{longtable}{|>{\centering\arraybackslash}m{1.4cm}|>{\centering\arraybackslash}m{0.8cm}|>{\centering\arraybackslash}m{2.15cm}|>{\centering\arraybackslash}m{2.3cm}|>{\centering\arraybackslash}m{2.8cm}|>{\centering\arraybackslash}m{1.3cm}|}
\caption{An Overview of Previous Research on DoE and MOO Approaches}
\label{tab:optimization_study} \\
\hline
\textbf{Ref} & \textbf{Year} & \textbf{Optimization Algorithm} & \textbf{Focus} & \textbf{Findings} & \textbf{Data-Driven/ Function}\\ \hline
\endfirsthead

\caption*{\textbf{Table \thetable} An Overview of Previous Research on DoE and MOO Approaches (continued)} \\ \hline
\textbf{Ref} & \textbf{Year} & \textbf{Optimization Algorithm} & \textbf{Focus} & \textbf{Findings} & \textbf{Data-Driven/ Function} \\ \hline
\endhead

Teja et al. & 2013 & Grey-Taguchi method & Optimizing surface roughness and MRR of stainless steel & Cutting speed having the most influence, followed by depth of cut and feed rate & Data Driven \\ \hline
Zhu et al. & 2016 & LHS, NSGA-II & Controlling precision forging & Low experimentation time to build a response surface approximation model & Function \\ \hline
Sakthivelu et al. & 2017 & Taguchi method and ANOVA & Developing optimum cutting conditions for minimum surface roughness and maximum MRR & Minimum surface roughness by controlling feed, depth of cut, and cutting speed, with depth of cut having the most significant influence on MRR & Data Driven \\ \hline
Lin et al. & 2018 & ISA using RBF, LHS, and AWS method & Reducing the computational cost of expensive black-box models & Well-distributed POF with weight reduction in electric bus body frame design & Function \\ \hline
Navid et al. & 2018 & Nelder-Mead with Sobol sequence and LHS & Optimizing Diesel engine & 7 RUNIDs by Sobol to find the solution, while in Latin Hypercube mode it takes 27 & Function \\ \hline
Zhao et al. & 2019 & RVEA, DBEA, $\epsilon$-MOEA, NSGA-III, and the IBEA & Evaluating the effectiveness of traditional algorithms & The superiority of the NSGA-III and RVEA algorithms & Function \\ \hline
Thakkar et al. & 2021 & MOO with CFD and RSA & Enhancing the performance of a centrifugal pump & Increasing head and efficiency of 9.154\% and 10.15\%, respectively & Function \\ \hline
Bulut et al. & 2021 & Combining LHS and GRA & Reducing pressure drop, maximizing maximum temperature, and convective heat transfer coefficient in batteries & A pressure drop reduction of up to 40.3\%, at the cost of an 11.3\% reduction in the convective heat transfer coefficient & Function \\ \hline
Chen et al. & 2022 & NSGA-II Genetic algorithm & Improving energy cost and efficiency for a prismatic battery module & Enhancement of the volume energy density by 9\% by developing a parallel liquid cooling system & Function \\ \hline
Mayda & 2022 & Kriging-GA with LHS & Optimizing the design of a robot arm & Stronger and lighter robot arm by high coefficients of determination and low RMS error & Data Driven \\ \hline
Alsharif et al. & 2023 & Ensemble ML models with orthogonal DoE and MOMRFO & Optimizing the geometrical arrangement of amorphous shading devices & Orthogonal structured dataset for EML model training due to low prediction errors and the TOPSIS method's higher payback periods & Function \\ \hline
Chang et al. & 2023 & LHS with the RS model and CGIDN & Increasing the productivity of plastic parts and optimizing injection molding parameters & Better performance of CGIDN in black box situations and improved accuracy through LHS, uniform, and orthogonal sampling& Data Driven \\ \hline
\end{longtable}

In the literature, optimization techniques have continually evolved to address MOO problems across a variety of industrial applications. Teja et al.\citep{teja2013multi} applied the Grey-Taguchi method to optimize surface roughness and material removal rate (MRR) for AISI 304 stainless steel during CNC milling, identifying cutting speed as the most influential factor.

Statistical DoE has proven to be a robust approach for enhancing process quality and efficiency. Zhu et al.\citep{zhu2016multi} utilized LHS and the NSGA-II algorithm to control the quality of precision forging, effectively reducing experimentation time by building a response surface approximation model. Sakthivelu et al.\citep{sakthivelu2017multi} investigated the machining characteristics of Aluminium Alloy using the Taguchi method to develop optimal cutting conditions for minimizing surface roughness, maximizing MRR, and finding the highest impact of depth of cut on MRR. Continuing into the later years, Lin et al.\citep{lin2018intelligent} proposed an Intelligent Sampling Approach (ISA) to minimize the computational costs of expensive MOO black-box simulation models. Navid et al.\citep{navid2018diesel} explored diesel engine optimization using the Nelder-Mead algorithm, with a distribution of starting points done by Sobol sequence and LHS, demonstrating the efficiency differences between Sobol and LHS modes in identifying optimal solutions. In 2019, Zhao et al.\citep{zhao2019comparative} ranked evolutionary many-objective (EMaO) algorithms based on optimization techniques, finding Reference Vector-guided Evolutionary Algorithm (RVEA) and Non-Dominated Sorting Genetic Algorithm III (NSGA-III) superior due to better performance, quality solutions, faster convergence, and Pareto front approximation.

The research momentum continued into the 2020s with significant contributions such as those from Thakkar et al.\citep{thakkar2021performance} who enhanced a sanitary centrifugal pump's performance using LHS, CFD, and a response surface approach, leading to noticeable improvements in head and efficiency. Concurrently, Bulut et al.\citep{bulut2021new} combined LHS and Grey Relational Analysis (GRA) to optimize battery design parameters, notably reducing pressure drop and optimizing temperature and convective heat transfer coefficients. In recent developments, Chen et al.\citep{chen2022multi} optimized a prismatic battery module using NSGA-II and a parallel liquid cooling system to enhance volume energy density by 9\%, showcasing the effectiveness of integrating advanced genetic algorithms with surrogate modeling. Similarly, Mayda\citep{mayda2022design} tackled the design optimization of a robot arm using the Kriging-genetic algorithm method, emphasizing the use of LHS in experimental design and MOGA in the optimization process.

In the subsequent year, Alsharif et al.\citep{alsharif2023multi} introduced innovative updates for optimizing shading devices, combining ensemble ML models with orthogonal DoE to enhance the geometrical arrangement of amorphous shading devices. Similarly, Chang et al.\citep{chang2023multi} employed LHS, the Response Surface Model, and the Constraint Generation Inverse Design Network (CGIDN) to optimize the injection molding process.

These studies collectively underscore the dynamic evolution of MOO strategies, demonstrating the critical role of statistical DoE and advanced metaheuristic algorithms in refining manufacturing processes and resolving complex optimization challenges across diverse industrial sectors.

\subsection{Development of Bayesian Optimization in Expensive MOO}
The concept of BO is not new, but it has attracted a lot of attention recently because of its incredibly flexible and successful implementation in a variety of expensive MOO scenarios. Table~\ref{tab:bayesian_optimization_study} provides a brief overview of the latest developments in BO and examples of its application to different multi-functional optimization problems.

%\begin{footnotesize} % Reduce the font size
%\begin{longtable}{|>{\centering\arraybackslash}m{1.8cm}|>{\centering\arraybackslash}m{1cm}|>{\centering\arraybackslash}m{2.6cm}|>{\centering\arraybackslash}m{3.1cm}|>{\centering\arraybackslash}m{3.45cm}|>{\centering\arraybackslash}m{1.9cm}|}
\begin{longtable}{|>{\centering\arraybackslash}m{1.4cm}|>{\centering\arraybackslash}m{0.7 cm}|>{\centering\arraybackslash}m{2.15 cm}|>{\centering\arraybackslash}m{2.4 cm}|>{\centering\arraybackslash}m{3 cm}|>{\centering\arraybackslash}m{1.3 cm}|}
\caption{An Overview of Prior Research Regarding Bayesian Optimization Framework}
\label{tab:bayesian_optimization_study} \\
\hline
\textbf{Ref} & \textbf{Year} & \textbf{Optimization Algorithm} & \textbf{Focus} & \textbf{Findings} & \textbf{Data Driven/ Function} \\ \hline
\endfirsthead

\caption*{\textbf{Table \thetable} An Overview of Prior Research Regarding Bayesian Optimization Framework (continued)} \\ \hline
\textbf{Ref} & \textbf{Year} & \textbf{Optimization Algorithm} & \textbf{Focus} & \textbf{Findings} & \textbf{Data-Driven/ Function} \\ \hline
\endhead

Emmerich et al. & 2011 & BO with hypervolume-based EI & Developing a computation algorithm of the hypervolume-based EI & Predictive distribution’s influence on the hypervolume-based EI & Function \\ \hline
Aboutaleb et al. & 2017 & FFF AM technique & Reducing geometric errors in AM & Reduction of the number of FFF trials by 20\% with the fewest errors & Data Driven \\ \hline
Solomou et al. & 2018 & BOED & Finding an experiment selection policy in MO materials discovery & Efficiency and consistency of the BOED framework & Data Driven \\ \hline
Talapatra et al. & 2018 & BMA within BO & Exploring the MDS and taking accounts of resource restrictions and model uncertainty & Auto-selection of the best features with limited initial data, eliminating the need for prior knowledge & Data Driven \\ \hline
Zerka et al. & 2021 & A distributed learning framework & Utilizing SL to exploit small sets of clinical and imaging data to train AI models & Superior performance compared to centralized learning for enhancing medical data privacy & Data Driven \\ \hline
Hanaoka & 2021 & Goal-oriented MOBO algorithm & Solving MO design difficulties & Acceleration of MO inverse design problems with minimal experiments & Data-Driven + Function \\ \hline
Botcha et al. & 2021 & Query by Committee active learning & Addressing alternatives for cost-effective experiments & Reduction of experimental costing by 50-65\% in the sequential approach & Data Driven \\ \hline
Hu et al. & 2023 & MOBO algorithm guided FEM & Speeding up the design of the Titania TPMS structure & Satisfying design requirements with a high modulus and permeability & Data Driven \\ \hline
Geng et al. & 2023 & A MOBO framework & Improving parameter settings in physical design tools & Exploration of high-quality parameter configurations & Function \\ \hline
Chepiga et al. & 2023 & Efficient MOBO Algorithm & Optimizing the parameters for laser powder bed fusion & Significantly less number of trials, saving time and materials & Data Driven \\ \hline
Ozaki et al. & 2024 & MOBO with Chebyshev scalarization-based utility function & A new acquisition function depending on the preference of the decision maker & Accelerating the optimization process and solving hyper-parameter optimization of  ML model & Function \\ \hline
Zhang et al. & 2024 & BO based on the qNEHVI & Optimization of Schotten-Baumann reaction in a flow & Optimal experimental conditions with fewer experiments and shorter calculation time & Data Driven \\ \hline
Khosravi et al. & 2024 & MOBOSL framework with qParEGO & Optimizing resource-intensive design parameters & Minimal data usage and introduction of a new performance metric (APHV) & Data Driven \\ \hline
\end{longtable}
%\end{footnotesize}

Emmerich et al.\citep{emmerich2011hypervolume} presented a computational algorithm of an acquisition function of BO known as Expected Improvement (EI), which was based on the hypervolume theory. Many years had gone by before scholars started to show interest again in using BO to significantly alter the manufacturing domains. In 2017, by improving the process parameter settings, Aboutaleb et al.\citep{aboutaleb2017multi} attempted to reduce geometric errors in items produced using a fused filament fabrication (FFF) AM technique. In comparison to the full factorial DOE approach, it reduced the number of FFF trials by 20\% with the fewest geometric errors, minimizing the amount of costly experimental trials in AM. In the subsequent year, Frazier\citep{frazier2018tutorial} explained BO using Gaussian process regression (GPR)~\citep{rasmussen:2006} and acquisition functions, discussing sophisticated techniques, BO tools, and future study areas. Solomou et al. \citep{Solomou2018} utilized Bayesian Optimal Experimental Design (BOED) in the same year to find targeted NiTi shape memory alloys using the Expected Hyper-Volume Improvement (EHVI) acquisition function and a GPR model. Talapatra et al.\citep{talapatra2018autonomous} also proposed a methodology at the same time to explore materials design space (MDS) while considering resource restrictions and model uncertainty. The technique blended Bayesian Model Averaging (BMA) with BO, autonomously and adaptively learning promising locations and models for exploration.

One of the critical reviews relating to this study was published by Greenhill et al.\citep{greenhill2020bayesian} in 2020, which explored the use of BO in experimental design as an alternative to commonly used DoE methods such as factorial designs, response surface approach, etc. The research on new frameworks of BO continued after 2020 and even intensified. In 2021, Zerka et al.\citep{zerka2021privacy} explored sequential learning's effectiveness in utilizing clinical and imaging data for AI model training, proposing a privacy-preserving distributed learning framework for Logistic Regression, Support Vector Machines, and Perceptron. At a similar time, a MOBO method was introduced by Hanaoka\citep{Hanaoka2021} that efficiently solved design challenges, reduced experimentation, and accelerated virtual multi-objective inverse design experiments over 1000 times compared to random sampling. Botcha et al.\citep{botcha2021efficient} also worked on sequential approaches in 2021 and showed that traditional ML methods, such as supervised learning techniques and neural networks, were impractical due to their high trial requirements. However, BO, with GP being its most widely used surrogate model, has faced substantial criticism for its inefficiency in modeling non-stationary and discontinuous functions over the years. To tackle the struggle of GP at discontinuities, Moustapha and Sudret\citep{moustapha2023learning} split the input space into regions with homogeneous behavior and fitted local GPs. This strategy involves clustering the data, categorizing fresh inputs to the appropriate region, and using local regression models. On the other hand, Snoek et al.\citep{snoek2014input} improved the performance of GP in handling non-stationary functions by applying a transformation to the input space. By learning bijective transformations of the input space using the Beta cumulative distribution function through warping inputs, GP models can enhance the performance of BO in non-stationary functions. Modifying the kernel function to account for non-stationarity can also help the GP to model varying behaviors across the input space \citep{sauer2023non}. It studies Bayesian Neural Networks (BNNs) as alternative surrogates, exploring different inference methodologies such as Hamiltonian Monte Carlo (HMC) and Deep Kernel Learning (DKL) to boost optimization performance.

As these advancements have significantly addressed the limitations of GP models, BO has gained increasing popularity, particularly in 2023, with a notable surge in research publications utilizing BO for single and multi-objective experimental design. A study was carried out at that time by Hu et al.\citep{Hu2023} to create a BO-guided Finite Element Method (FEM) analysis process to speed up the design of Titania TPMS designs. Geng et al.\citep{Geng2023} proposed an information gain-based MOBO framework to tune the parameter settings for a physical design tool. Following that, the laser powder bed fusion (L-PBF) process was optimized using a new algorithm named Diversity-Guided Efficient Multi-objective Optimization (DGEMO) by Chepiga et al.\citep{chepiga2023process}. It created a perfect processing window for producing SS 316L efficiently and inexpensively, reducing trial numbers and saving time and materials. Later, in 2024, Ozaki et al.\citep{ozaki2024multi} proposed a BO approach for MOO problems with expensive objective functions. They used a human-in-the-loop approach, a Chebyshev scalarization-based utility function, and an active learning strategy to accelerate the optimization of benchmark functions. At the same time, Khosravi et al.\citep{khosravi2024data} introduced a data-driven BO framework along with sequential learning to analyze complicated systems with conflicting goals. Moreover, a brand-new metric for assessing MOO techniques was suggested. So, it can be concluded that in recent years, BO has grabbed the attention of researchers in many fields, especially in the costly manufacturing optimization processes.

\section{Methodology}
\label{Methodology}
This section describes the terminologies used in the study, the workflow of the models, the detailed process of our proposed framework, the performance metrics, and finally, an analysis of the numerical data set.
\subsection{Preliminaries}
The section introduces the basic concepts of MOO and the key components of the BO, such as the surrogate model and acquisition function.
\subsubsection{Multi-Objective Optimization (MOO)}
The formulation of a general MOO problem with N objectives to be optimized is as follows \citep{alvarado2020multi}:
\begin{equation}
\begin{aligned}
    &\text{Max or min} && F(x) = [f_1 (x), f_2 (x), \dots, f_N (x)] \\
    &\text{subject to} && p(x) = [p_1 (x), p_2 (x), \dots, p_i (x)] \leq 0 \\
    &&& q(x) = [q_1 (x), q_2 (x), \dots, q_j (x)] = 0,
\end{aligned}
\end{equation}
where  $x = [x_1, x_2, \ldots, x_N]$ represents a vector of decision variables, where $x_i$ denotes the $i^{th}$ variable. The objective function $f_N(x)$ represents the $N^{th}$ objective to be optimized. Additionally, $p_i(x)$ and $q_j(x)$ correspond to the $i^{th}$ inequality constraint and $j^{th}$ equality constraint, respectively. It is often challenging to find a single solution that maximizes all objectives while satisfying all constraints simultaneously. This difficulty leads to the concept of Pareto optimality, which is commonly utilized in analyzing objective vectors. 

\subsubsection{Pareto Front}
Pareto dominance and Pareto optimality are widely used to compare potential solutions to MO problems. Non-dominated solutions, also known as Pareto optimal solutions, represent compromises or trade-offs between objectives, where improving one objective cannot be achieved without deteriorating another. Consequently, they represent a potentially infinite number of solutions \citep{ngatchou2005pareto}. A solution $x^*$ is Pareto optimal if, for any other solution $\bar{x}$ in the solution space $S$, $f_k(x^*) \geq f_k(\bar{x})$ (for maximization) or $f_k(x^*) \leq f_k(\bar{x})$ (for minimization) for $k = 1, 2, 3, \ldots, N$, and for at least one $k$, $f_k(x^*) > f_k(\bar{x})$ (for maximization) or  $f_k(x^*) < f_k(\bar{x})$ (for minimization) \citep{Natarajan2020}. The Pareto front comprises all Pareto optimal solutions. The mathematical expression for the PF is as follows:
%In the case of minimization problem, a solution $x^*$ is Pareto optimal if, for any other solution $\bar{x}$ in the solution space $S$, $f_k(x^*) \leq f_k(\bar{x})$ for $k = 1, 2, 3, \ldots, N$, and for at least one $k$, $f_k(x^*) < f_k(\bar{x})$.
\begin{equation}
\begin{aligned}
     PF = \{ x^* \in S \mid & \nexists \bar{x} \in S, \bar{x} \neq x^*, I * f_k(x^*) \geq I * f_k(\bar{x}) \, \forall k = 1,2,\ldots,N \land \exists k, \\ & I * f_k(x^*) > I * f_k(\bar{x}) \}  %&\text{[Maximization Problem]}
\end{aligned}
\end{equation}
where \( I = 1 \) in maximization problems and \( I = -1 \) in minimization problems.
%egin{equation}
%\begin{aligned}
    %&& PF = \{ x^* \in S \mid \nexists \bar{x} \in S, \bar{x} \neq x^*, f_k(x^*) \leq f_k(\bar{x}) \, \forall k = 1,2,\ldots,N \land \exists k, f_k(x^*) < f_k(\bar{x}) \}  &\text{[Minimization Problem]}
%\end{aligned}
%\end{equation}
\subsubsection{Bayesian optimization (BO)}
\begin{figure}[!ht]
    \centering
    \includegraphics[height= 2.45 in,keepaspectratio]{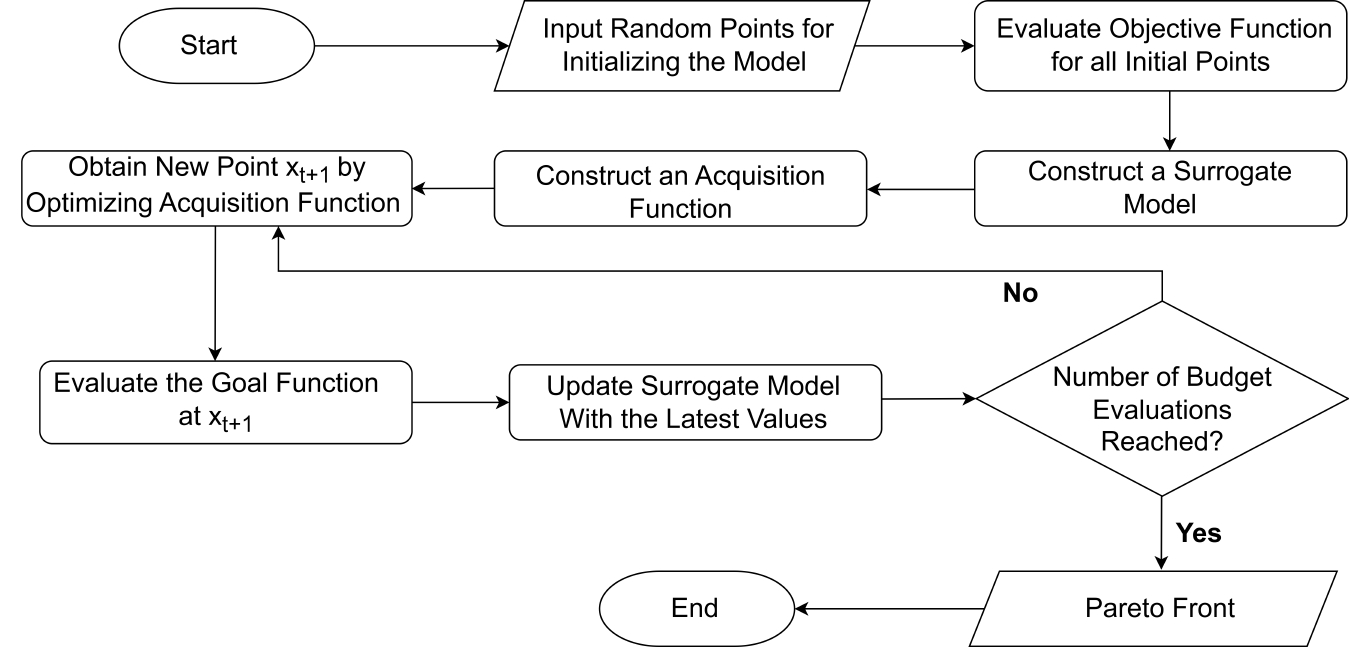}
    \caption{Bayesian Optimization Workflow}
    \label{bayesian workflow}
\end{figure}

The working principle of BO is outlined as follows:
\begin{itemize}
    \item The process begins by defining the goal function f(x), which is typically modeled using a GP~\citep{rasmussen:2006}. This modeling facilitates the creation of a surrogate model for the objective function.
    \item The prior distribution is initially populated with random values of input features and the objective function is evaluated at these points.
    \item Before proceeding, the probability of improvement is assessed at each position within the search space. This assessment is facilitated by the development of an acquisition function. Common acquisition functions include Lower Confidence Bound (LCB), Upper Confidence Bound (UCB), Probability of Improvement (PI), and Expected Improvement (EI) \citep{frazier2018tutorial}.
    \item The acquisition function plays a crucial role in selecting the next point for experimentation, striving to balance the exploration of new, potentially promising regions with the exploitation of areas known to contain local optima. Each data acquired and incorporated into the model updates it, shifting the prior distribution to a posterior distribution, thereby integrating new insights into the system's behavior.
\end{itemize}
This iterative process continues to refine the understanding of the function's landscape, optimizing the search for the best outcomes by efficiently navigating the balance between exploration and exploitation. The BO workflow is highlighted in Fig.~\ref{bayesian workflow}.

\subsubsection{Surrogate Model}
The surrogate model plays a crucial role in simulating complex design systems, providing an efficient approximation of the underlying response function. Among commonly used surrogate models, Random Forest Regression is an effective interpolator that delivers highly accurate predictions within the vicinity of training data \citep{yang2017mixed, breiman2001random}. However, its performance deteriorates significantly when extrapolating beyond the observed data, making it unreliable for modeling regions with sparse or distant training samples \citep{morozovska2017ferroionic}. 

An alternative to Random Forest Regression is the GP model, which offers a probabilistic framework with built-in uncertainty quantification. While GP is well-suited for low-dimensional problems, it struggles with scalability in high-dimensional and computationally expensive domains. Neural networks (NNs) provide a more advanced surrogate modeling approach, excelling in handling high-dimensional and complex datasets where GPs may become computationally intractable \citep{ansari2023large}. However, NNs introduce significant computational overhead, particularly in small datasets, and are prone to overfitting and sensitivity to initialization. The computational cost reductions achieved by BO over traditional DoE methods could be negated or even exacerbated when a resource-intensive NN-based surrogate model is employed. Instead of using a single NN, an ensemble of multiple NNs can be trained as a surrogate model. This approach enhances generalization, mitigates overfitting, and provides more robust uncertainty estimation compared to a single NN \citep{lei2021bayesian}. However, for small or low-dimensional datasets, this method remains computationally inefficient. Furthermore, while typical NNs lack inherent uncertainty quantification, GPs offer built-in predictive uncertainty through their probabilistic formulation.

In this study, we adopt certain assumptions regarding the GP model to facilitate computational efficiency. First, we assume stationarity, implying that the statistical properties of the process remain invariant across the input space. Consequently, the covariance between outputs depends solely on the relative positions of inputs rather than their absolute locations. Additionally, we assume the dataset is continuous and differentiable, necessitating the use of a kernel function that enforces smoothness in function approximation. In cases where these assumptions are violated, we recommend input warping or the use of non-stationary kernels to effectively model non-stationary functions \citep{sauer2023non, snoek2014input}.

GP encapsulates the opinions held regarding the connections between input and output variables. These assumptions are based on a predetermined dataset, $D_{1:t} = \{(x_1, y_1), (x_2, y_2), \dots, (x_t, y_t)\}$, where $x_t$ stands for the input variables and $y_t$ stands for the output variables at time $t$. We have multiple $y_t$'s in the multi-objective problem.

A GP model can be characterized by its mean $\mu(x)$ and the covariance $k(x,x')$:
\begin{equation}
    f(x) \sim \text{GP}(\mu(x), k(x,x')).
\end{equation}
Here, \( k(x,x') \) is frequently referred to as the ``kernel”. It is anticipated that if two points are near each other, then so will their corresponding process outputs. The degree of proximity varies with the distance between the points. One of the often-used options for the covariance function is the Squared Exponential (SE) function, sometimes referred to as the Radial Basis Function (RBF)~\citep{greenhill2020bayesian}, which can be expressed as:
\begin{equation}
k(x,x') = \exp \left( -\frac{1}{2\theta^2} \|x-x'\|^2 \right),
\end{equation} where \(\theta\) is the characteristic length scale, controlling the width of the radial basis functions and \(||x-x'||\) represents the Euclidean distance between the input vectors \( x\) and \( x'\), measuring the distance or dissimilarity between the two points in the input space.
The model can be summarized as follows, and an experimental setup includes introducing a term for normally distributed noise: $\varepsilon \sim \mathcal{N}(0,\sigma_{\text{Noise}}^2)$. 
\begin{equation}
y = f(x) + \varepsilon,
\end{equation}
where \( y \) is the process output. GPR can be used to predict the goal function's value at time \( t+1 \) for any location \( x \). According to \citep{greenhill2020bayesian}, this leads to a normal distribution with a mean \( \mu_t(x) \) and an uncertainty of \( \sigma_t(x) \). The posterior distribution of the following equation, which reflects the design space as it is now understood, can be incrementally modified with new experimental data \citep{khosravi2024data}.
\begin{equation}
    P(f_{t+1} \mid D_{1:t}, x) = \mathcal{N}(\mu_t(x), \sigma_t^2(x)),
\end{equation}
\begin{equation}
\begin{aligned}
    \text{where} \quad \mu_t(x) &= k^T [K + \sigma_{\text{noise}}^2 I]^{-1} y_{1:t}, \\
    \sigma_t(x) &= k(x,x) - k^T [K + \sigma_{\text{noise}}^2 I]^{-1} k, \\
     \quad \text{ \&} \quad k &= [k(x,x_1), k(x,x_2), \ldots, k(x,x_t)].   
\end{aligned}
\end{equation}

\( K \) is the kernel matrix of already sampled designs \( x_t \) which is defined as follows:
\begin{equation}
K = \left[ \begin{array}{ccc}
k(x, x_1) & \cdots & k(x, x_t) \\
\vdots & \ddots & \vdots \\
k(x_t, x_1) & \cdots & k(x_t, x_t)
\end{array} \right].
\end{equation}

Although we have analytical formulae, achieving precise inference in GP regression has a computational complexity of $\mathcal{O}(n^3)$, where $n$ denotes the number of observations. This significant cost derives from the necessity to invert the covariance matrix. In reality, the Cholesky decomposition can be precomputed and stored, decreasing the cost of subsequent predictions to $\mathcal{O}(n^2)$. Nonetheless, each time kernel hyperparameters are typically re-optimized at every iteration, requiring the recalculation of the Cholesky decomposition, thereby maintaining the cubic complexity for model updates. This computational overhead makes precise inference infeasible for large datasets or when a high number of function evaluations is required. Consequently, numerous approximation techniques, such as sparse GPs, inducing points, and low-rank approximations, have been developed to mitigate these computational costs \citep{shahriari2015taking}.

\subsubsection{Acquisition Function}
An acquisition function is built to reflect the ideal conditions for the upcoming experiment. The \(\mu(x)\) and \(\sigma(x)\) of the GP model serve as the main sources of the acquisition functions, which are easy to compute. The acquisition function, whose global maximizer is employed as the next experimental setting, enables a trade-off between exploitation (sampling where the objective mean is high) and exploration (sampling where the uncertainty is large) \citep{greenhill2020bayesian}.

\subsection{Traditional Space-filling Design of Experiment (DoE) Methods}
In order to understand complex systems and improve their performance, experiments are planned and carried out systematically using the DoE method. DoE enables a thorough investigation of the design space, resulting in a better comprehension of the behavior of the system and the identification of optimal solutions. This is accomplished by carefully choosing the experimental settings and levels for each input variable \citep{weissman2015design}. The capacity of DoE to offer an organized and effective framework for experimental design is key to its efficacy. It permits the creation of a small number of well-chosen samples that yield detailed insights into the system's response. DoE facilitates the finding of Pareto-optimal solutions, which are the optimum trade-offs between competing objectives, by guaranteeing a representative coverage of the design space \citep{mayda2022design}. 

\begin{figure}[!ht]
    \centering
    \includegraphics[height=2.68 in,keepaspectratio]{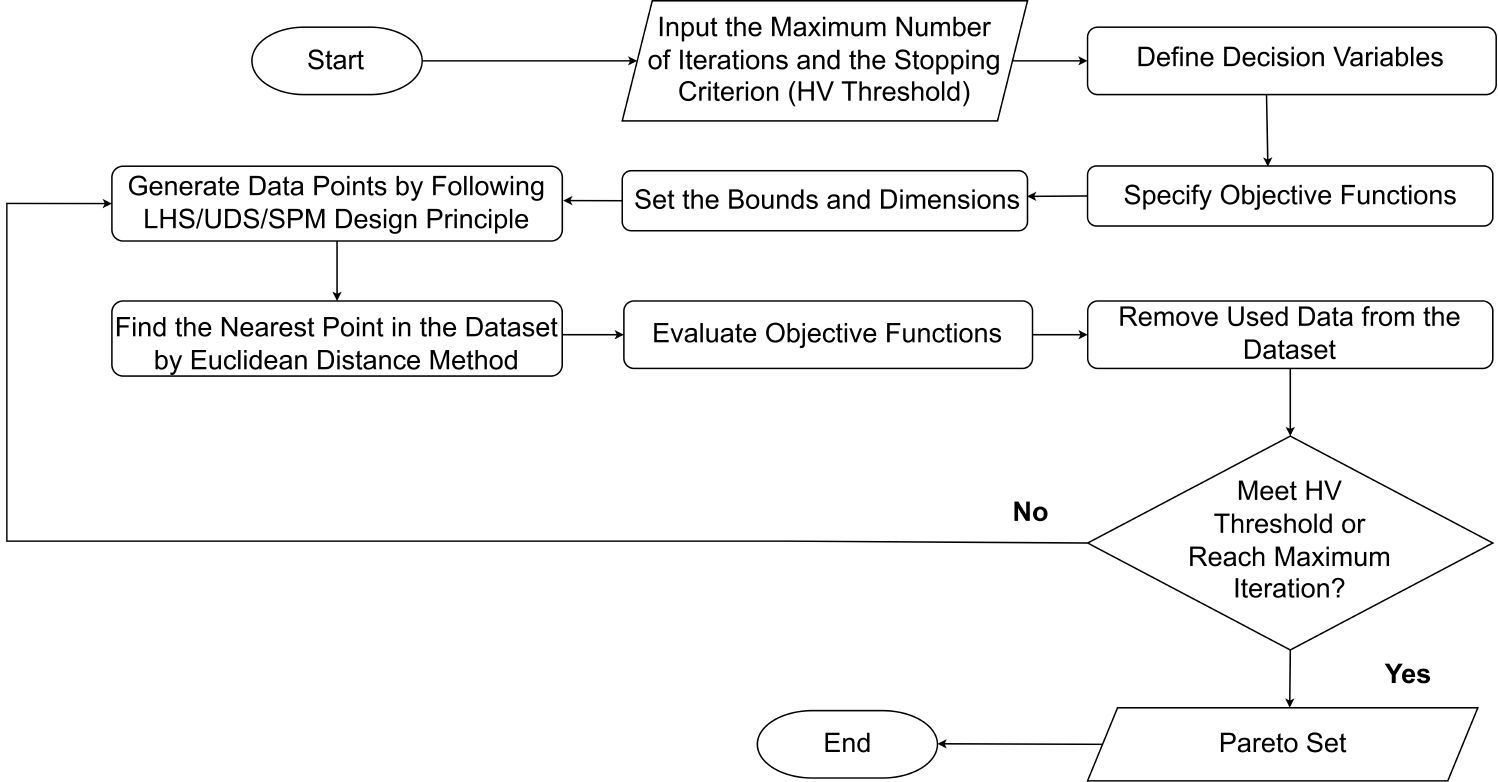}
    \caption{Workflow of Space-filling Design of Experiment }
    \label{doe workflow}
\end{figure}

DoE, however, also has certain drawbacks. DoE's assumption of linearity and additivity of effects may not apply to complex systems, which is one of its limitations. In MOO, where the interactions between objectives and variables are complex, DoE necessitates a good knowledge of the issue and the selection of relevant variables and levels. DoE also calls for a large number of tests to be run, which in reality may be both time-consuming and expensive \citep{miranda2020advantages}. In this study, 3 methods of space-filling DoE: Uniform Design Sampling, Latin Hypercube Sampling, and Sphere Packing Method are utilized. The general workflow of the space-filling DoE is illustrated in Fig.~\ref{doe workflow}.

\subsubsection{Uniform Design Sampling (UDS)}
Uniform Sampling is a deterministic sampling technique that is one sort of Quasi-Monte Carlo (MC) approach. It fills in the space on the experimental region uniformly by searching for its representational points or rep-points \citep{jing2005efficient}. The resulting estimator may have a relatively high variance, which would be detrimental to the underlying optimization to converge \citep{zhao2015stochastic}.

Since UDS is simple to apply, it is frequently employed in situations where the distribution of the variables is unknown or assumed. In uniform sampling, there is an equal chance of selection for every point in the experimental space. There is no discernible pattern; the samples are dispersed evenly throughout the whole parameter space. Every point is chosen at random, without regard to the others \citep{fang2000uniform}. Uniform sampling distributes points randomly throughout the whole parameter space without any pattern. 

\subsubsection{Latin Hypercube Sampling (LHS)}
LHS is a statistical sampling method used to explore the parameter space of a design issue quickly. It is especially helpful when there are several criteria or considerations to take into account. LHS creates a stratified matrix of samples by dividing the range of each parameter into equally likely intervals and ensuring that only one sample is obtained from each interval. In comparison to random sampling, LHS offers a more uniform and accurate coverage of the parameter space \citep{cioppa2007efficient}. LHS's main goal is to produce samples as varied as possible while lowering sampling uncertainty \citep{helton2003latin}. To explore a greater variety of parameter combinations and gain a deeper grasp of the design space, LHS is helpful \citep{Stein1987}.

LHS does have certain drawbacks, too. LHS's assumption of a linear relationship between the input parameters and the response variables is one of its limitations \citep{liefvendahl2006study}. Additionally, LHS may have local optima problems and does not guarantee global optimum. Since LHS works best with continuous variables, it is also inappropriate for issues involving discrete or categorical variables \citep{loh1996latin}.

\subsubsection{Sphere Packing Method (SPM)}
The DOE methodologies are essential for addressing intricate optimization challenges, particularly in extremal planar geometry. Sphere packing, a key method in this field, involves the arrangement of circles within a given space to maximize efficiency \citep{audet2007extremal, lubachevsky2004dense}. One notable challenge is determining the maximum radius $r$ of $n$ identical circles that fit inside a square. The challenge is to find the maximum radius of $n$ identical circles, ensuring the shortest path between their centers fits into a unit square \citep{hifi2009literature}. One way to conceptualize this is as a continuous global optimization problem where the decision variables' optimal level, $r$ must be determined as follows:

\begin{equation}
\begin{aligned}
& \text{Maximize } \quad r  \\
& \text{subject to }\quad r \leq x_i \leq 1 - r \quad \text{; } \quad i \in I \\
& \quad\quad\quad\quad \quad r \leq y_i \leq 1 - r \quad \text{; } \quad i \in I \\
& \quad\quad\quad\quad \sqrt{(x_i - x_j)^2 + (y_i - y_j)^2} \geq 2r \quad \text{; } \quad 1 \leq i < j \leq n,
\end{aligned}
\end{equation}

where \((x_i, y_i)\), \(i \in I\{1, 2, 3, \ldots, n\)\} denotes the coordinates of the center of circle \(i\), and \(\sqrt{(x_i - x_j)^2 + (y_i - y_j)^2}\) is the Euclidean distance between the centers of circles \(i\) and \(j\) \citep{hifi2009literature}.

\subsection{Performance Metrics}
This section introduces five performance metrics employed in the study: Generational Distance (GD), Inverted Generational Distance (IGD), Hypervolume (HV), Proportional Hypervolume (PHV), and Data Usage (D).
\subsubsection{Generational Distance (GD)}
The GD measure calculates the distance between a solution and the Pareto front and may be used to assess an algorithm's performance \citep{van1999multiobjective}. Assume that the points discovered by the algorithm are used to generate the set of objective vectors, \( \textbf{A} = \{ a_1, a_2, a_3, \ldots, a_{\lvert \textbf{A} \rvert} \} \), and the Pareto front is represented by \( P = \{ p_1, p_2, p_3, \ldots, p_{\lvert P \rvert} \} \). The GD value can be found using the equation below:
\begin{equation}
GD(\textbf{A}) = \frac{1}{\lvert \textbf{A} \rvert} \left( \sum_{i=1}^{\lvert \textbf{A} \rvert} d_i^n \right)^{\frac{1}{n}},
\end{equation}
where \( d_i \) is the Euclidean distance from \( a_i \) and its closest point of reference in \( P \). This represents the average distance between any point in \( \textbf{A} \) and the closest point on the Pareto front. In this equation, \( n \) is always equal to 2 \citep{ishibuchi2015modified}.

\subsubsection{Inverted Generational Distance (IGD)}
IGD measures the average distance between the solutions in the true Pareto front and the nearest solutions in the approximation set, as opposed to GD's calculation of the average distance between the solutions in the approximation set (generated by the algorithm) and the nearest solutions in the true Pareto front \citep{ishibuchi2018reference}. Because it offers a more sensitive and thorough evaluation of the approximation set's quality by taking into account both convergence and diversity throughout the whole Pareto front, IGD is typically chosen over GD. As a result, IGD becomes a more reliable and insightful indicator for assessing how well MOO algorithms perform. IGD calculates the distance 
between a point in \(P\) and its nearest point in \(\textbf{A}\) \citep{ishibuchi2015modified}. The IGD equation is as follows:
\begin{equation}
IGD(\textbf{A}) = \frac{1}{\lvert P \rvert} \left( \sum_{i=1}^{\lvert P \rvert} \hat{d}_i^n \right)^{\frac{1}{n}},
\end{equation}
where \( \hat{d}_i \) is the Euclidean distance (\( n = 2 \)) from \( p_i \) to its closest point in \( \textbf{A} \).

\subsubsection{Hypervolume (HV)}
In MOO, the term ``hypervolume" is used to quantify how much a collection of solutions dominates or covers the objective space. The hypervolume is the volume of the objective space that is bounded by the reference point and dominated by the solutions. The reference point acts as a standard against which to compare the solutions' coverage \citep{ishibuchi2018specify}. The mathematical expression of the hypervolume metric is demonstrated in Eq.~\ref{eq:hv} as Hypervolume \( HV(F, R) \) for a Pareto front \( F \) and a reference point \( R \) in the \( N \)-dimensional objective space \( \mathbb{R}^N \) \citep{Manoj2023}.
\begin{equation}
\label{eq:hv}
    HV(F,R) = \phi_N \left( \bigcup_{i=1}^{\lvert F \rvert} HR_i \right),
\end{equation}
where \( \phi_N \) is the \( N \)-dimensional Lebesgue Measure and \( HR_i \) is the hyper-rectangle formed by the \( i^{th}\) Pareto point and \( R \) as the vertices. To ensure that the solutions cover as much of the objective space as possible, the hypervolume must be maximized \citep{Manoj2023}. Analyzing hypervolume allows for comparison and evaluation of optimization techniques, providing a quantitative method to assess efficiency and performance in MOO.

\subsubsection{Proportional Hypervolume (PHV)}
 In MOO, PHV is used to assess how well a Pareto front approximation performs. The ratio of the HV covered by the generated Pareto Points to the HV created by true Pareto points is measured in this metric. PHV represents the variation between the actual Pareto points and generated ones \citep{aboutaleb2017multi}.
 \begin{equation}
     PHV = \frac{\text{HV(Resulted Pareto Points)}}{\text{HV(True Pareto Points)}}.
 \end{equation}
 PHV is a measurement that falls between $[0, 1]$, and in a perfect world, PHV should equal to 1.

\subsubsection{Data Usage (D)}
Data utilization is one of the key performance metrics for evaluating algorithms involving costly experiments. The measurement of this indicator tells us how much data is required to attain a threshold of PHV or HV. Mathematically, the following equation is the formula for determining this metric:
\begin{equation}
    D = \frac{\text{Total Data Points Used to Reach at the Expected Pareto Front}}{\text{Highest Number of Data Points Available}}.
\end{equation}

\subsection{Proposed Bayesian Multi-objective Sequential Decision-Making (BMSDM) Framework}
This section presents a data-driven, efficient MOBO paradigm to solve costly black-box MOO issues. We utilize Quasi Expected Hypervolume Improvement (qEHVI) as the acquisition function and the GP as the surrogate model of our proposed framework. This section first introduces the qEHVI, followed by the full model formulation of BMSDM.

\subsubsection{Quasi Expected Hypervolume Improvement (qEHVI)}
Traditional DOE approaches have laid the groundwork for more advanced techniques in MOO. One such method is the Expected Hypervolume Improvement (EHVI) \citep{emmerich2005single}, which has been a cornerstone for optimizing expensive and complex systems. EHVI is particularly valuable as it quantifies the expected improvement in the hypervolume of the objective space, guiding the selection of new candidate points to balance exploration and exploitation effectively.

An extension of EHVI to the parallel, limited evaluation context can be defined as quasi-expected hypervolume improvement (qEHVI). We can optimize and propose the next best site because of the parallelism added in this version of EHVI. This lessens the computational effort and needs less time when costly evaluations in batches are desired \citep{biswas2021multi}. Batch BO is often employed for expensive black-box functions, where the computational cost of batch approaches is small relative to the function evaluation cost. In such cases, the computational overhead of batch techniques does not greatly impair optimization performance. However, when function evaluations are affordable, global optimization approaches like DIRECT or multi-start Newton methods may be more suitable. In such instances, a computationally costly batch BO strategy may not be the best solution \citep{nguyen2018practical}. The combined EHVI of q new candidate points is computed exactly up to the MC integration error by qEHVI. Auto-differentiation makes it possible to generate precise gradients of the MC estimator, in contrast to earlier EHVI formulations that depend on gradient-free acquisition optimization.

The Hypervolume measure \citep{deb2011multi}, a well-known metric, serves as the foundation for qEHVI, which assesses the Pareto front's convergence quality. After selecting the initial values of input features and finding corresponding output features, the GP model is built primarily. Then, a reference point is established based on search space knowledge, which very closely approximates the targeted Pareto curve \citep{cho2023multi}. This is already illustrated mathematically in Eq.~\ref{eq:hv}. BO is employed by the theory of maximizing the Hypervolume Improvement (HVI) on the Pareto front to identify the subsequent sample point. The following equation presents how to evaluate \(HVI\) owing to \( x_{new} \):
\begin{equation}
HVI(x_{new}) = HV(P_{new},R) - HV(P,R),
\end{equation}
where \( P_{{new}} \) is the Pareto front that contains the objectives of \( x_{new} \). The expected value of the improvement in \(HV\) for adding \( x_{new} \) is denoted as \(EHVI( x_{new} \)).
\begin{equation}
    EHVI(x_{new}) = \int HVI(x_{new}) \cdot PDF(x) \, dx,
\end{equation}
where \(PDF\) is an evaluation of a multivariate independent normal distribution model. EHVI has a substantial level of computational complexity. In this respect, the qEHVI, which calculates EHVI in parallel, can be used to increase optimization efficiency in real-world issues.
Let \( S_i \) be the space that is dominated by the \( i^{th} \) point in the \( q \) sample points but not by \( P \). The following equation presents the expression for \(HVI\) for a group of \( q \) points (\( qHVI \)):

\begin{equation}
    qHVI(\{X_i\}_{i=1}^{q}) = \phi_K \left( \bigcup_{i=1}^{q} S_i \right).
\end{equation}
The union of \( S_i \) is a region dominated by \( q \) new points, and Lebesgue integration \( \phi_K \left( \bigcup_{i=1}^{q} S_i \right) \) is a joint \(HVI\) composed of \( q \) new points \citep{cho2023multi}. The surrogate model does not generate point estimates but rather prediction distributions of the objectives due to its stochastic nature. The HVI turns out to be stochastic as the HVI depends on the objective function value at \( x_{\text{new}} \), which is stochastic too. It follows that the acquisition function should be designed with the expectation of \(HVI\) over the posterior distribution of objectives, as shown in Eq.~\ref{eq:HVI} :
\begin{equation}
\label{eq:HVI}
    qEHVI(\{X_i\}_{i=1}^{q}) = \int_{-\infty}^{\infty} (qHVI) \, df,
\end{equation}
where \( f \) is the joint posterior of the objective functions. The calculation of \( qEHVI \) does not have an analytical form when the objectives are correlated, which is very usual in MOO problems. An MC approximation of \( qEHVI \) with \( N \) samples is taken uniformly using the Sobol sampling method, as given in Eq.~\ref{eq:ehvi}:
\begin{equation}
\label{eq:ehvi}
qEHVI(\{X_i\}_{i=1}^{q}) \approx \frac{1}{N} \sum_{t=1}^{N} (qHVI).
\end{equation}
The volume of \( 2^q - 1 \) hyper-rectangles must be determined using \( qEHVI \) for every \( K \) hyper-rectangle and \( N \) MC samples. The formulation of qEHVI is extremely parallelizable and, despite its computational expense, might attain constant time complexity with limitless processor cores. By exploiting recent GPU power and computing accurate gradients, qEHVI optimization beats existing state-of-the-art approaches in numerous practical circumstances\citep{daulton2020differentiable}.

\subsubsection{Model Formulation of Bayesian Multi-objective Sequential Decision-Making (BMSDM) Framework}
Our framework was designed utilizing Python and the BoTorch package for efficient optimization. The implementation and experiments were conducted in Google Colab, a free, cloud-based Jupyter notebook environment that allows access to CPUs, GPUs, and TPUs. While Colab enables hardware acceleration, no external GPU was used for this project. For local execution, a 7th-generation Intel Core i5 processor with 8GB of RAM was utilized. This arrangement relied primarily on CPU computing capacity, necessitating careful computational management due to hardware restrictions. The BO framework successfully conducted optimization tasks, but the computing cost was notable—90 iterations required around 30 minutes. This amounts to an average time of 20 seconds in every iteration, showing the influence of hardware limits on execution time.

\begin{figure}[!ht]
    \centering    \includegraphics[width=4.25in,keepaspectratio]{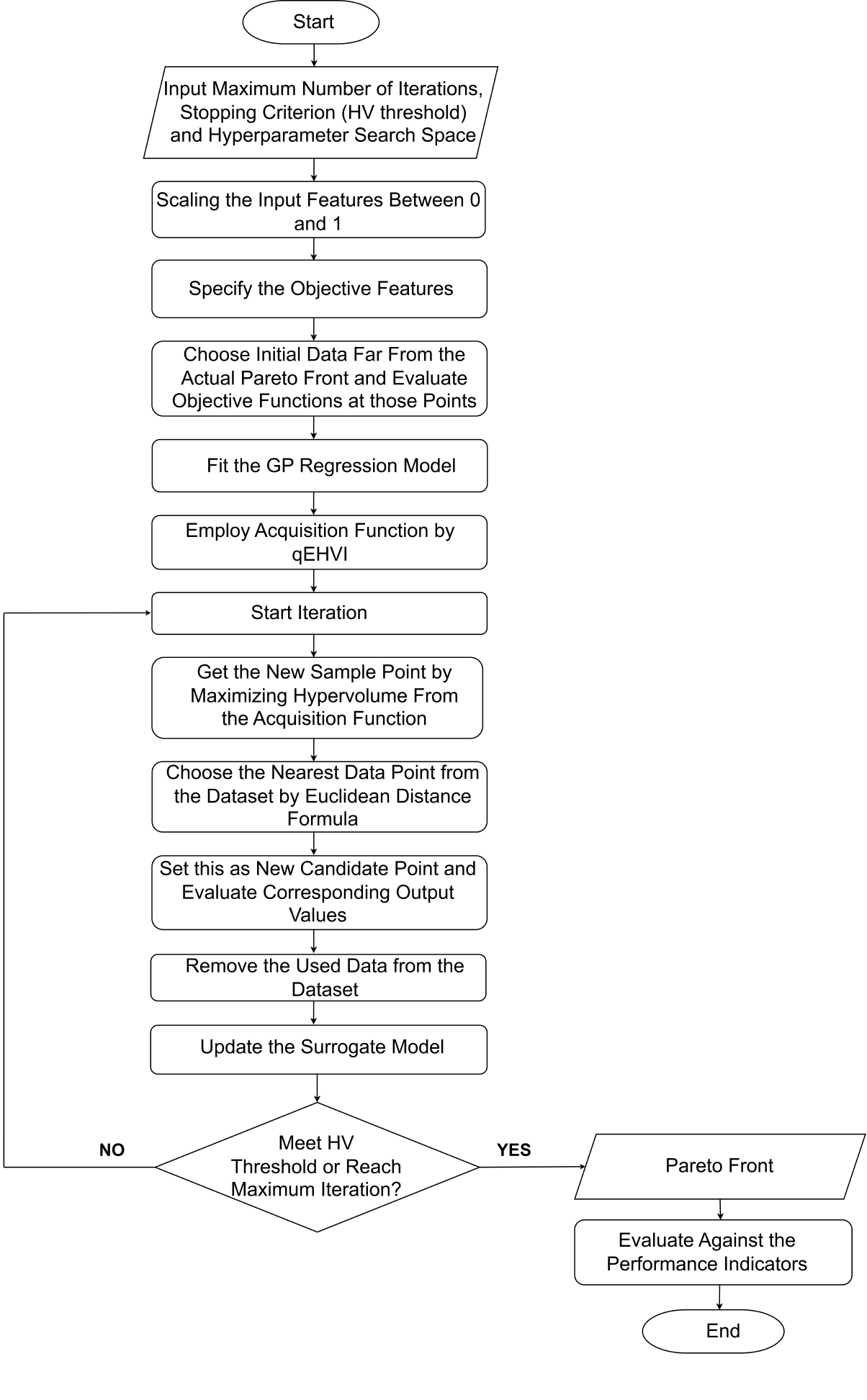}
    \caption{Workflow of BMSDM Framework}
    \label{qEHVI}
\end{figure}

The workflow of the proposed method is summarized in Fig.~\ref{qEHVI}. Based on this Figure, to guide the optimization process effectively, the method initiates by identifying and normalizing the input features within a range of 0 to 1. Subsequently, essential parameters such as the hyperparameter search space, HV threshold (0.95 for the max-max scenario and 0.94 for the max-min scenario), and the upper limit on iterations need configuration. It is crucial to clearly define the objectives for optimization, particularly in an MO problem where multiple target features are involved. The selection of input variables is driven by the specific context of the problem and the goals of the optimization. Following the selection of inputs, a GP model is fitted to an initial dataset, which is chosen far from the optimal points. The qEHVI acquisition function is employed to identify the most promising candidate for the next sample point, aiming to maximize the hypervolume in the search space. In deploying the qEHVI, the selection of a reference point representing the least favorable solution is crucial. In a max-max scenario, this would be the lowest point for each objective feature. In contrast, in a scenario with competing objectives, the reference point is the least desirable value for the feature to be maximized and the highest for the feature to be minimized.

The process then enters its main iterative phase. Given that the model is validated on a pre-existing dataset, there's a possibility that the points suggested by the acquisition function might not exist in the dataset. This necessitates a method to approximate the closest feasible point for practical implementation. To address this, the Euclidean distance between each suggested point and all existing points in the dataset is calculated, which then updates the GP model as the new sample. This data point is subsequently excluded from future selections to avoid repetition. The iteration continues until either the maximum number of iterations is reached or the desired hypervolume threshold is met. With each cycle, a new candidate is integrated into the surrogate model. This iterative process seeks a balance between leveraging new information for optimal solutions and exploring novel areas that could yield superior outcomes. The final step involves identifying the Pareto front, representing a set of non-dominated solutions where any improvement in one objective would lead to a compromise in another. The algorithm's effectiveness is then evaluated using various performance metrics.

In summary, the BMSDM methodology is shown in Algorithm~\ref{alg:algorithm} and comprises:
\begin{itemize}
    \item Initial steps involving input feature establishment and scaling, selection of initial data far away from the true Pareto front, and fitting a GP model to this data.
    \item Generation of new candidate points through the qEHVI function, followed by Euclidean distance calculations to determine the closest sample point.
    \item Continuation of the optimization process until reaching the HV threshold or the maximum iteration count, culminating in the identification of the Pareto front, a collection of non-dominated sample points.
\end{itemize}
\FloatBarrier
\begin{algorithm}[h]
  \caption{\textit{Bayesian Multi-objective Sequential Decision-Making (BMSDM) Framework}}
  \label{alg:algorithm}
  \begin{algorithmic}
    \Statex 
        \textbf{Inputs:} Dataset $D$ with $l$ records, black-box objectives: $f(x)=(f_1(x), f_2(x), \ldots, f_m(x))$; Max iterations limit: $n\_iter$; Initial sample count: $n\_start$; Batch size: $b$; Search space dimensionality: $\text{num\_dims}$; HV threshold for max-max scenario: $\text{hv\_max\_max}$; HV threshold for max-min scenario: $\text{hv\_max\_min}$; distance array: $d_m$.
    \Statex
        \textbf{Output:} The set of non-dominated points $(X, Y)$ in Pareto front.
    \State
        \textbf{\textit{// Step 1: Initialization Phase}}
    \State
        01: Select $n\_start$ initial points from $D$, ensuring they are not close to the actual Pareto front, and evaluate $f(X)$ at these points
    \State 
        02: Normalize inputs in $D$ from 0 to 1
    \State
        03: Fit an initial GP model on $(X, Y)$
    \State
        \textbf{\textit{// Step 2: Optimization Loop}}
    \State
        04: Set iteration counter $i = 1$ and current HV $= 0$
    %\State
        %05: Set current HV = 0
    \State
        05: {\textbf{While}{$(i \leq n\_iter)$ and current HV $< \text{hv\_max\_max}$ (or $\text{hv\_max\_min}$)}} \textbf{Do}:    
    \State
        \textbf{\textit{// Step 3: Candidate Generation Step}}
    \State
        \hspace{\algorithmicindent} 06: Generate new candidate point $x_{\text{new}}$ with qEHVI
    \State
        \hspace{\algorithmicindent} 07: \textbf{For}{$j = 1$ to $l$ (each point $D[j]$ in dataset $D$)} \textbf{Do}:
    \State
        \hspace{\algorithmicindent \hspace{\algorithmicindent}} 08: Compute Euclidean distance $d_m[j] = \|x_{\text{new}} - D[j]\|$
    \State
        \hspace{\algorithmicindent}09: \textbf{End For}
    \State
        \textbf{\textit{// Step 4: Selecting Next Sample Point}}
    \State
        \hspace{\algorithmicindent} 10: min\_index = Find index of the minimum value in $d_m$
    \State
        \hspace {\algorithmicindent} 11: $x_{\text{next}} = D[\text{min\_index}]$
    \State
        \textbf{\textit{// Step 5: Model Update Phase}}
    \State
        \hspace{\algorithmicindent}12: Add the new sample to the set $X$
    \State
        \hspace{\algorithmicindent}13: Re-train the GP and qEHVI with updated data
    \State
        \hspace{\algorithmicindent}14: Update current HV based on new Pareto front
    \State
        \hspace{\algorithmicindent}15: Increment $i$
    \State
        16. \textbf{End While}
    \State
        \textbf{\textit{// Step 6: Final Output}}
    \State
        17: \textbf{Return} the Pareto front of $(X, Y)$
    \State
        18: \textbf{End}
  \end{algorithmic}
\end{algorithm}
\FloatBarrier

\section{Result Analysis}
\label{Result}
This section introduces the numerical dataset used in the study, followed by an evaluation of the proposed model against three space-filling DoE methods using multiple performance metrics. The primary goal is to generate a high-quality Pareto front with minimal data points. A comprehensive analysis is provided based on the results from the four methods discussed in the previous section. To assess model stability, boxplots from 25 computational runs are also presented. The final subsection compares the performance of our proposed approach with two state-of-the-art MOO methods from distinct algorithmic families.

\subsection{Numerical Data Set Analysis}
In this study, we utilize a numerical data set that explores the early transition metal carbides and nitrides, which are known as MAX or M$_\text{n+1}$AX$_\text{n}$ phase \citep{barsoum2000mn+}. According to the periodic table, M stands for a transition metal, A for group IV and VA elements, and X for either carbon or nitrogen \citep{makino2000estimation}. The model framework includes 402 MAX phases in the design space. Similar to the other sequential learning systems, our study assumes that the MDS of the MAX phases is known beforehand, which includes 15 features \citep{talapatra2018autonomous}. In this data, C and m are empirical constants that connect the material's bulk modulus to its constituent parts \citep{makino2000estimation}. Notably, the amount of valence electrons, C$_\text{v}$, is linked to the character of bonding. This is recognized as a sign of a phase's stability \citep{guo2011effect, karthikeyan2015role}. The description of all the features of the dataset is given in Table~\ref{tab:features}.

\begin{table}[htbp]
  \centering
  \caption{Dataset Features with Description}
   \label{tab:features}
    \begin{tabular}{|>{\centering\arraybackslash}m{3.2 cm}|>{\centering\arraybackslash}m{9cm}|}
    \hline
    \textbf{Feature(s)} & \textbf{Description} \\ \hline
    C and m & Empirical constants \\
    \hline
    $\text{C}_{\text{v}}$ & Valence electron concentration \\
    \hline
    e/a & Electron-to-atom ratio \\
    \hline
        a, b and c & Lattice parameters \\
    \hline
    Z & Atomic number \\
    \hline
    Ductility & Material's ability to deform under stress without fracturing \\
    \hline
    Type & The specific type of MAX phase material \\
    \hline
    $\text{I}_{\text{dist}}$ & Interatomic distance \\
    \hline
     $\text{Col}_{\text{M}}$,  $\text{Col}_{\text{A}}$, and  $\text{Col}_{\text{X}}$ & The groups according to the periodic table of the M, A \& X elements \\
    \hline
    X(M), X(A), and X(X) & Electronegativity of M, A and X elements \\
    \hline
    APF & Atomic packing factor \\
    \hline
    radius & Average atomic radius \\
    \hline
    volume & Volume/atom \\
    \hline
    G & Shear Modulus \\
    \hline
    K & Bulk Modulus \\
    \hline
    E & Modulus of Elasticity \\
    \hline
    \end{tabular}%
\end{table}

The MAX phase dataset adheres to strict criteria to ensure the robust evaluation of the framework. Properties such as elastic modulus, stability, and thermal conductivity exhibit variability across experimental and computational sources, making proper normalization essential before integrating them into the framework. Stability is particularly critical, as an unstable MAX phase may lead to shifts in fundamental descriptors such as atomic radius, electronegativity, and valence electron concentration, introducing inconsistencies in the dataset. Since the framework's evaluation relies on comparing its outcomes with the real Pareto front, maintaining a comprehensive and accurate dataset is imperative. Erroneous or missing data can result in misaligned Pareto fronts, compromising the reliability of the evaluation. The selection of seven input features and two output features is grounded in physical insights and experimentally observed relationships, ensuring that the framework effectively captures meaningful material property patterns \citep{khosravi2024data}. Additionally, some features exhibit highly specific distributions, lack randomness (e.g., interatomic distance, electron-to-atom ratio, ductility, volume), or contain missing or redundant data, making them unsuitable as input or output features. By adhering to these restrictions, the framework ensures reliable optimization and provides a valid assessment of performance in material discovery.
\begin{table}[!ht]
\centering
\caption{The Primary Parameters of the Proposed BMSDM Approach}
\label{tab:parameters}
\begin{tabular}{|>{\centering\arraybackslash}m{2.4 cm}|>{\centering\arraybackslash}m{7cm}|>{\centering\arraybackslash}m{3.8cm}|}
\hline
\textbf{Parameter} & \textbf{Description} & \textbf{Value(s)} \\ \hline
\textit{b} & Batch size & 5 \\ \hline
\textit{n\_iter} & Maximum number of iterations (Resource Restriction Scenario) & 90 \\ \hline
\textit{NUM\_RESTARTS} & Quantity of chance restarts necessary to optimize an acquisition function & 10 \\ \hline
\textit{RAW\_SAMPLES} & Number of random samples to generate from a given search space & 402 \\ \hline
\textit{bounds} & Lower and upper bounds of a search space (a tensor) & [0,1] \\ \hline
\textit{MC\_SAMPLES} & Number of Monte Carlo samples to use when estimating the expected improvement & 32 \\ \hline
\textit{n\_start} & Number of initial samples & 30, 10 (Resource Restriction Scenario) \\ \hline
\textit{hv} & HV threshold & 0.95 (max-max scenario) and 0.94 (max-min scenario) \\ \hline
\textit{ref\_pt} & Reference point for calculating the hypervolume (a tensor) & [0,0] (max-max scenario) and [50,170] (max-min scenario) \\ \hline
\end{tabular}
\end{table}

\subsection{Defining the Initial Model Parameters}
The model incorporates several parameters (Batch size, Quantity of chance restarts necessary to optimize qEHVI, Number of MC samples to use when estimating the expected improvement) that are derived from the Botorch module \citep{daulton2020differentiable} by default and largely used during the optimization phase. Before starting the framework, input features were normalized to a standard bound between 0 and 1, ensuring uniform scaling and better optimization performance. For hypervolume calculation, the reference point is chosen as the worst point (a tensor) in the Pareto front, based on the optimization scenario:
\\
\textbf{Max-Max Scenario:} The reference point is (0,0), as it represents the minimum values for both output features.
\\
\textbf{Max-Min Scenario:} The reference point is (50,170), where Objective 1 takes its minimum value, and Objective 2 takes its maximum value—ensuring accurate hypervolume assessment.

However, other parameters, such as \textit{hv, n\_start}, and \textit{n\_iter}, are determined based on the characteristics of the dataset and the specific situations, like resource restriction scenario or full cycle experimentation, presented in this study. In our dataset, each feature comprises 402 data points collected under unrestricted resource conditions. One of our evaluation approaches involves leveraging the entire dataset to assess the Pareto front and compare the performance metrics of our model against existing frameworks. However, real-world applications often face resource constraints, such as machine failures, limited availability of time and materials, or operational disruptions, restricting data collection. To simulate such practical constraints, we introduce a resource-restriction scenario, where only 25\% of the available resources are accessible at a given time, limiting the dataset to 100 data points. Within this subset, 10 data points are randomly selected for real experiments, while the framework strategically determines 90 sequential candidate points for further experimentation to optimize the output features. This approach ensures that the model is evaluated under realistic constraints, demonstrating its robustness and adaptability in situations where complete datasets are unattainable. Additionally, it highlights the framework's efficiency in sequential decision-making, addressing challenges inherent to data-scarce environments. The remaining experimental parameters are detailed in Table~\ref{tab:parameters}.

\subsection{Result Analysis of the BMSDM, UDS, LHS, and SPM}
This research incorporates 7 input features and 2 output features defining the material design space. The input features include atomic packing factor (APF), average atomic radius (rad), valence electron concentration (C$_\text{v}$), position in the periodic table of elements A and X (Col$_\text{A}$ and Col$_\text{X}$, respectively), and the electronegativity of elements X and A. The model evaluates performance under two distinct scenarios: Scenario 1 features both objectives at their maximum (max-max). In contrast, Scenario 2 involves one objective at its maximum and the other at its minimum (max-min). In Scenario 1, the focus is on identifying materials with the highest bulk modulus (K) and shear modulus (G). In Scenario 2, the aim is to find materials with the highest bulk modulus but the lowest shear modulus. Materials that require low shear modulus and high bulk modulus are typically chosen for their robustness and resistance to deformation. Conversely, materials intended for high stiffness applications are sought for their high values in both bulk and shear moduli \citep{talapatra2018autonomous}.

\subsubsection{ Max-Max Scenario}
All of the models have been examined first to check how many data points are needed to achieve the actual Pareto front. No restriction of data points is imposed here and the performance of the models is judged based on the value of D metric.

\begin{figure}[!ht]
  \centering
  \begin{tabular}{cc}
    \includegraphics[width=0.35\linewidth, keepaspectratio]{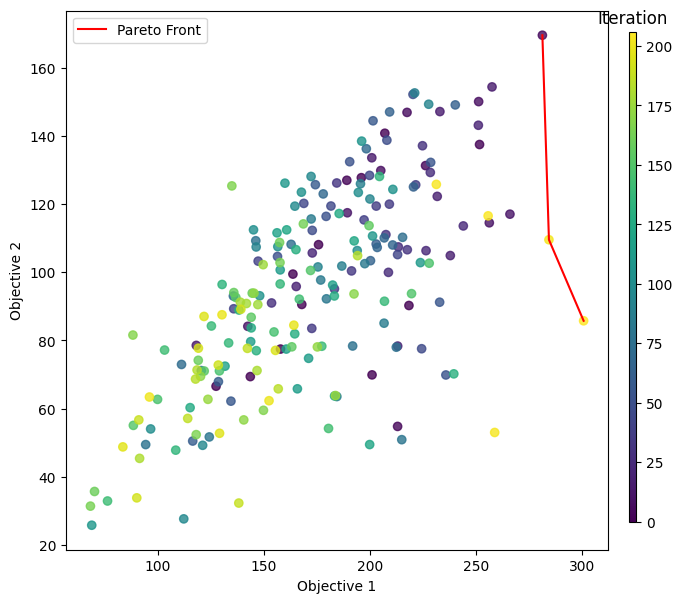} & \includegraphics[width=0.35\linewidth, keepaspectratio]{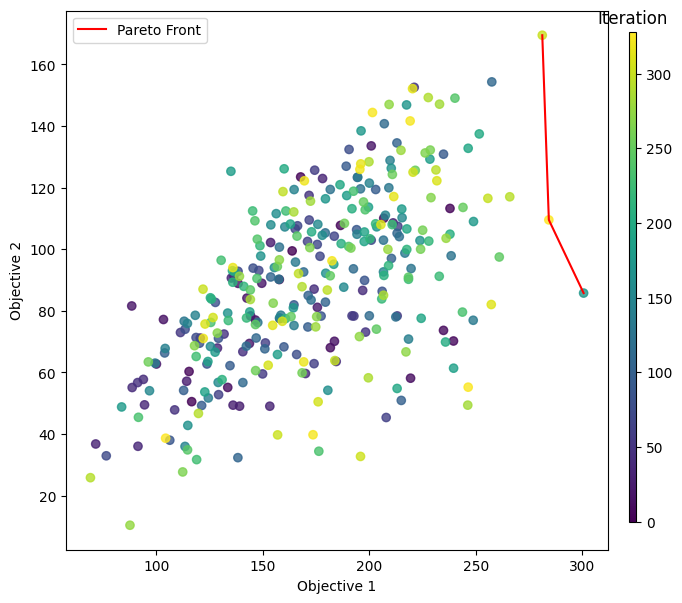} \\
    BMSDM & LHS \\
    Data Points Needed: 53.73\%  & Data Points Needed: 81.84\% \\
    \includegraphics[width=0.35\linewidth, keepaspectratio]{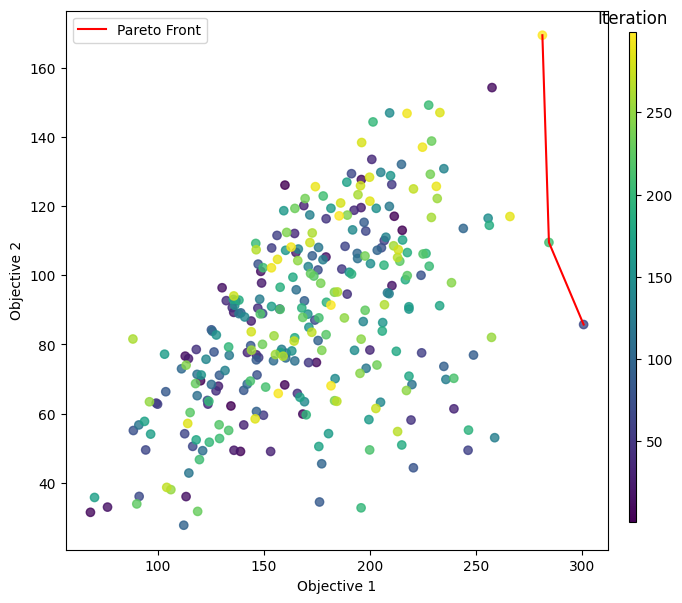} & \includegraphics[width=0.35\linewidth, keepaspectratio]{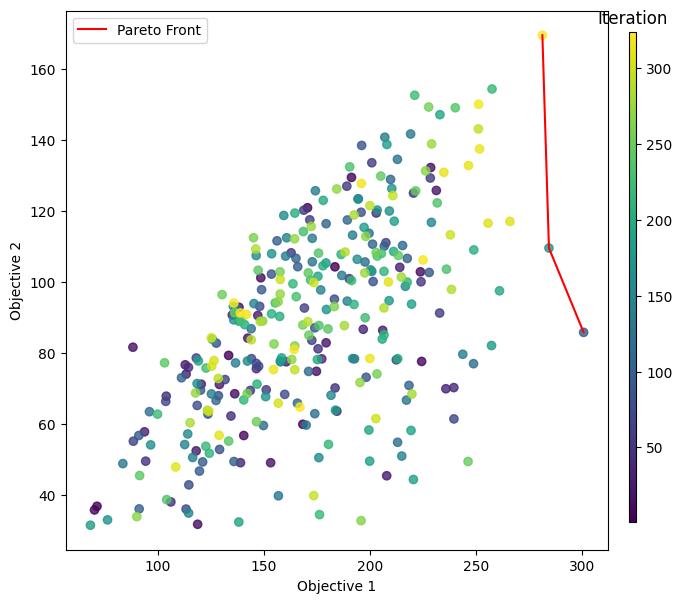} \\
    UDS & SPM \\
    Data Points Needed: 74.38\% & Data Points Needed: 80.6\% 
  \end{tabular}
  \caption{Comprehensive Comparison Among the Models with respect to Data Usage (D)}
  \label{tab:fig8}
\end{figure}

Based on Fig.~\ref{tab:fig8}, we can see that BMSDM achieves a true Pareto front with a significant amount of less data, proving the efficiency of the algorithm. In comparison to UDS, which ranked second, BMSDM requires about 90 fewer data points to get the same outcome. The performance of SPM and LHS are almost similar, though LHS performs worst according to this metric having more than 81\% data points. Upon thoroughly examining the BMSDM method, we can determine the number of iterations required to meet specific performance criteria by splitting the total iteration required into four parts: the first iteration, 25\% iteration, 50\% iteration (half cycle), and 100\% iteration (full cycle).
\begin{figure}[!ht]
  \centering
  \begin{tabular}{cc}
    \includegraphics[width=0.45\linewidth, keepaspectratio]{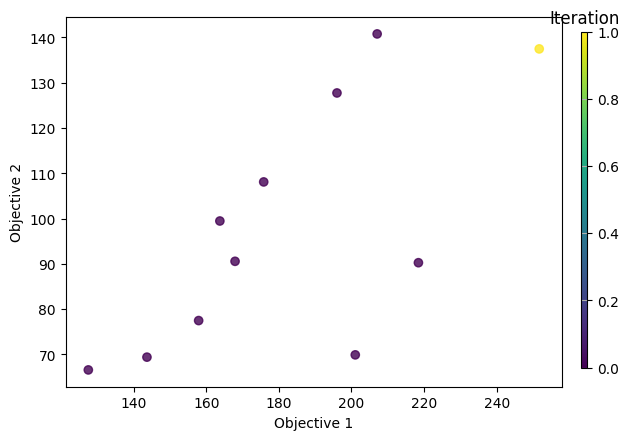} & \includegraphics[width=0.45\linewidth, keepaspectratio]{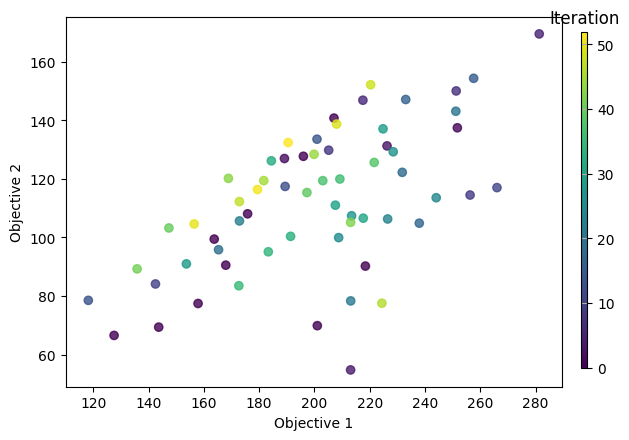} \\
    1$^\text{st}$ Iteration: & 25\% Iteration: \\
    GD: 63.97 \quad IGD: 61.3 & GD: 0.0047 \quad IGD: 0.0047 \\
    HV: 0.69 \quad PHV: 71.39\% & HV: 0.93 \quad PHV: 96.48\% \\
    \includegraphics[width=0.45\linewidth, keepaspectratio]{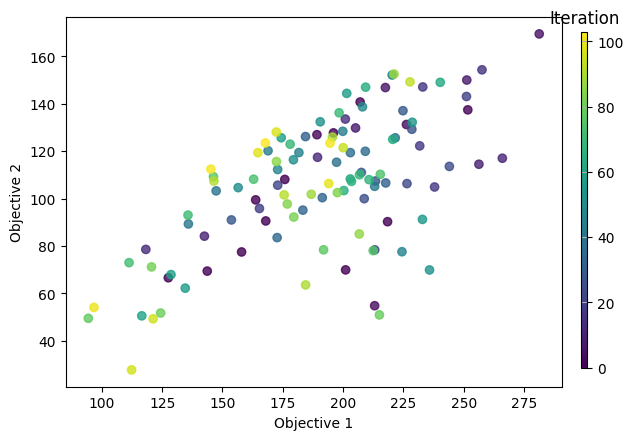} & \includegraphics[width=0.45\linewidth, keepaspectratio]{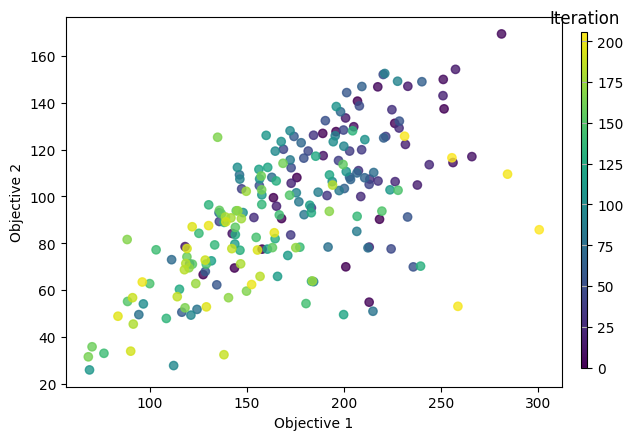} \\
    50\% Iteration: & 100\% Iteration: \\
    GD: 0.0047 \quad IGD: 0.0047 & GD: 0.0053 \quad IGD: 0.0052 \\
    HV: 0.93 \quad PHV: 96.48\% & HV: 0.97 \quad PHV: 100\% 
  \end{tabular}
  \caption{Performance Analysis of BMSDM After Different Numbers of Iterations}
  \label{tab:fig9}
\end{figure}

Fig.~\ref{tab:fig9} indicates that a minimal amount of data is sufficient to achieve 95\% of the PHV value. Increasing the iteration to 50\% does not yield improved outcomes. However, after a full 100\% iteration, the model captures all real Pareto points. Analysis of the full dataset (402 data points) revealed that beyond 25\% of iterations, further computational costs did not yield significant improvements in performance (Fig.~\ref{tab:fig9}). Increasing the sample size unnecessarily would lead to excessive computational overhead without a proportional benefit in optimization effectiveness. In our resource restriction scenario, we imitate real-world constraints by limiting available data. When only 25\% of resources are accessible, we operate with 100 data points—selecting 10 randomly (far from the Pareto points) for initial training, while the framework strategically selects the remaining 90 sequential candidate points to optimize the output features. This approach ensures that our model remains robust and effective even when complete data collection is impractical, reinforcing the validity of our evaluation methodology. Moreover, this study demonstrates that the algorithm successfully reaches its objectives using an initial sample located outside of the Pareto front.

Given the dataset's 7 input features and their domain-specific ranges, 90 sequential iterations were determined to provide an optimal trade-off—ensuring diverse coverage of the feature space while maintaining computational feasibility. This choice was based on observed diminishing returns, where additional iterations beyond 90 offered minimal performance gains compared to the increased computational cost.

\begin{figure}[!ht]
  \centering
  \begin{tabular}{cc}
    \includegraphics[width=0.45\linewidth]{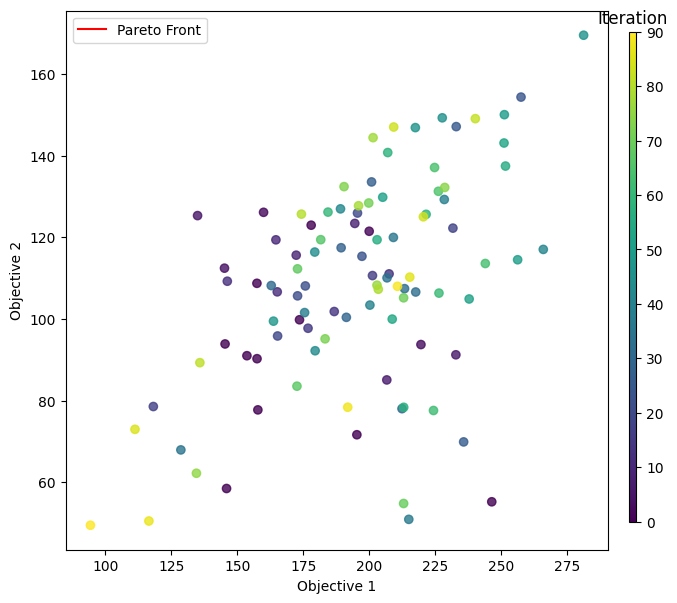} & \includegraphics[width=0.45\linewidth]{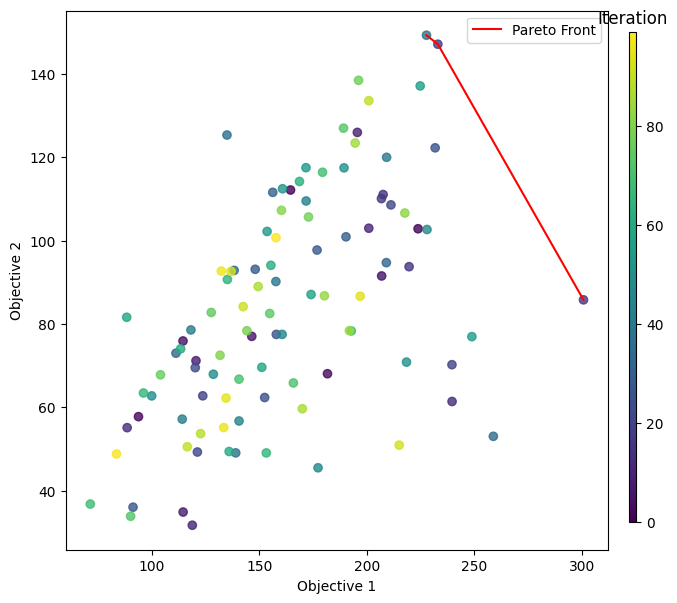} \\
    BMSDM & LHS \\
    GD: 0.0047 \quad IGD: 0.0047 & GD: 45.13 \quad IGD: 36.83 \\
    HV: 0.93 \quad PHV: 96.48\% & HV: 0.79 \quad PHV: 82.10\% \\
    \includegraphics[width=0.45\linewidth]{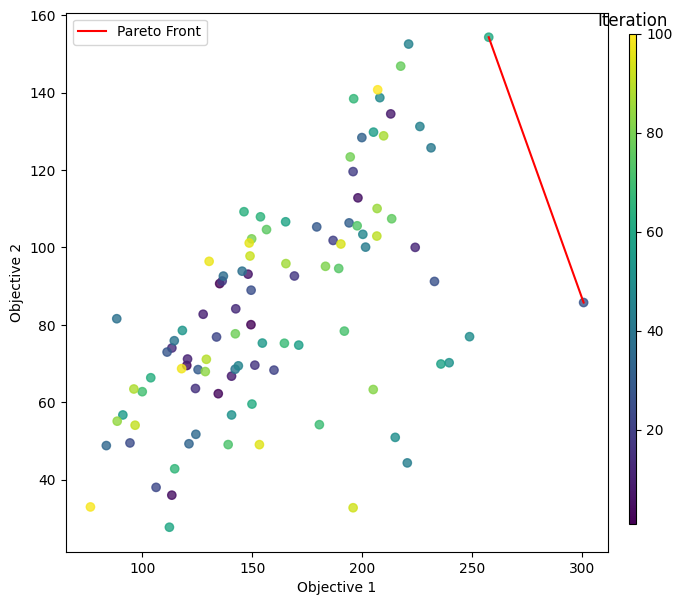} & \includegraphics[width=0.45\linewidth]{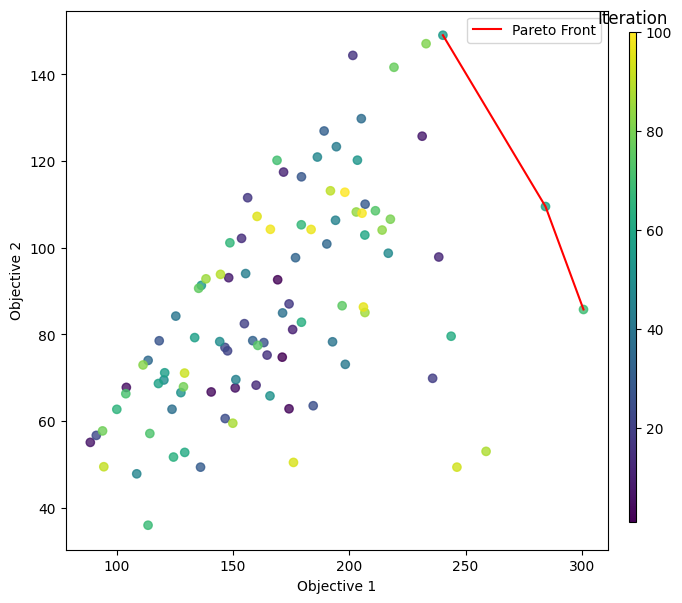} \\
    UDS & SPM \\
    GD: 19.9 \quad IGD: 14.07 & GD: 26.47 \quad IGD: 15.29 \\
    HV: 0.85 \quad PHV: 87.93\% & HV: 0.82 \quad PHV: 85.01\%
  \end{tabular}
  \caption{Comprehensive Comparison Among the Models Using Performance Metrics}
  \label{tab:fig10}
\end{figure}

Fig.~\ref{tab:fig10} reveals that BMSDM excels across all metrics, achieving optimal GD and IGD values. SPM records a greater GD value than UDS, although both perform similarly in terms of IGD. Compared to SPM, UDS performs better on the other performance metrics. SPM stands third in this scenario, where LHS shows the weakest performance among the methods, with HV below 80\% and higher GD and IGD values than its counterparts. However, due to the inherent randomness of these methods, drawing conclusions from a single run may not be reliable. To provide a more dependable evaluation of the GD, IGD, HV, and PHV values, and to demonstrate the models' stability, twenty-five runs are conducted. The performance of each model is depicted in a boxplot, which includes up to 100 data points.
\begin{figure}[!ht]
  \centering
  \begin{tabular}{cc}
    \includegraphics[width=0.45\linewidth, keepaspectratio]{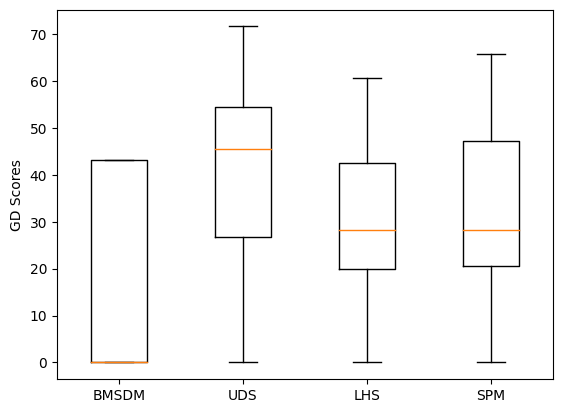} & \includegraphics[width=0.45\linewidth, keepaspectratio]{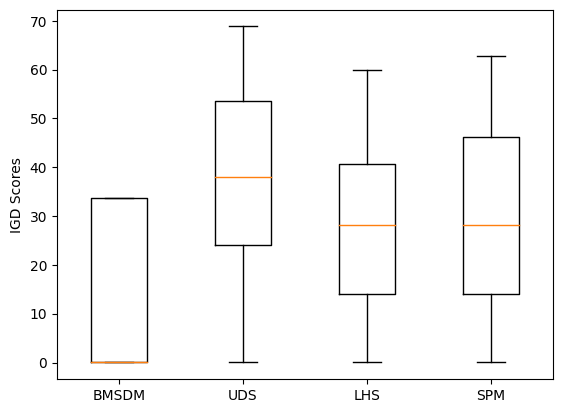} \\
    GD & IGD \\
    \includegraphics[width=0.45\linewidth, keepaspectratio]{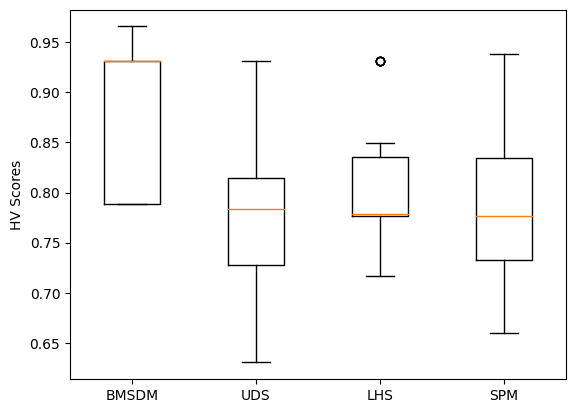} & \includegraphics[width=0.45\linewidth, keepaspectratio]{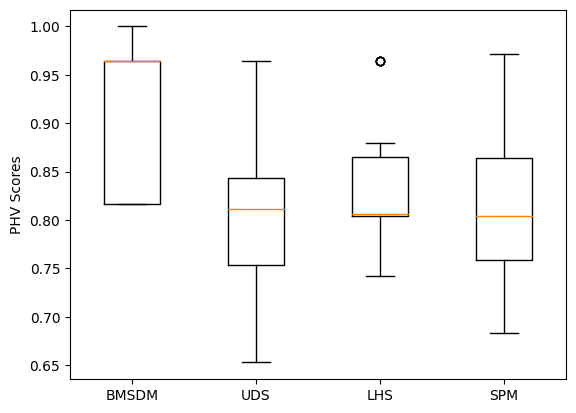} \\
    HV & PHV
  \end{tabular}
  \caption{Stability of the Models Based on Different Performance Metrics}
  \label{tab:fig11}
\end{figure}

It is evident from Fig.~\ref{tab:fig11} that BMSDM outperforms the other three methods on every performance criterion. Even though the performance of all the space-filling techniques is nearly identical, UDS is the second-best algorithm in terms of PHV value, followed by LHS, and SPM, which has the poorest performance. While BMSDM attains a PHV value of over 95\% after a relatively small number of iterations, the traditional approaches only yield about 80\% of the PHV value. The median GD and IGD of BMSDM is nearly zero, which is the most desired value when considering the GD and IGD values. 

\subsubsection{Max-Min Scenario}
In the context of decision-making with competing objectives (max-min scenarios), identical methodologies are applied across all models. To evaluate the number of data points required for each model to accurately converge to the true Pareto front, a comprehensive examination without data limitations is conducted for each model. The efficacy of these models is assessed using the metric `D', which serves as a performance indicator.
\begin{figure}[!ht]
  \centering
  \begin{tabular}{cc}
    \includegraphics[width=0.45\linewidth]{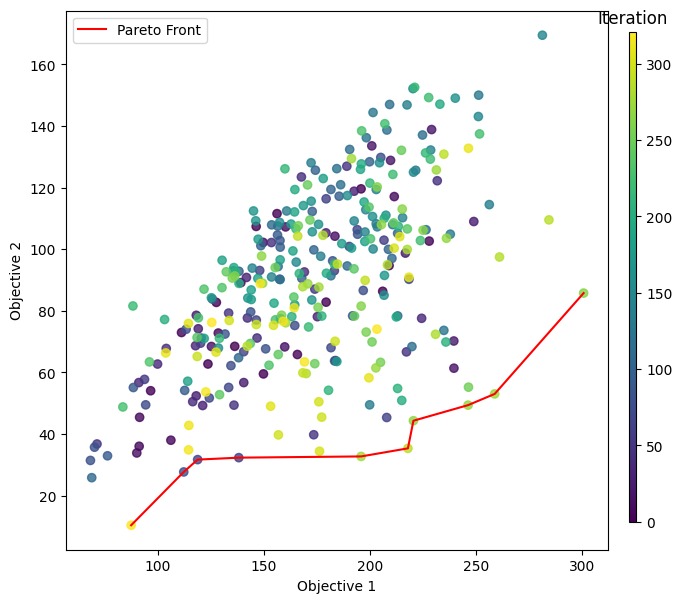} & \includegraphics[width=0.45\linewidth]{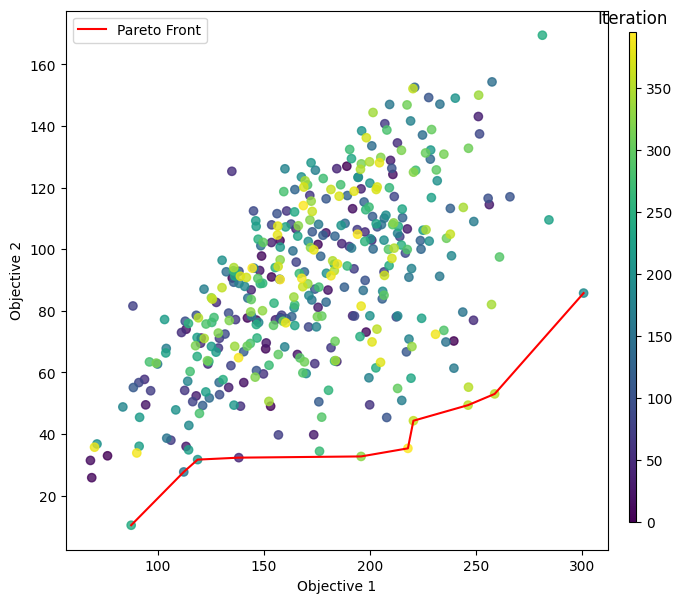} \\
    BMSDM & LHS \\
    Data Points Needed: 82.33\% & Data Points Needed: 98.51\% \\
    \includegraphics[width=0.45\linewidth]{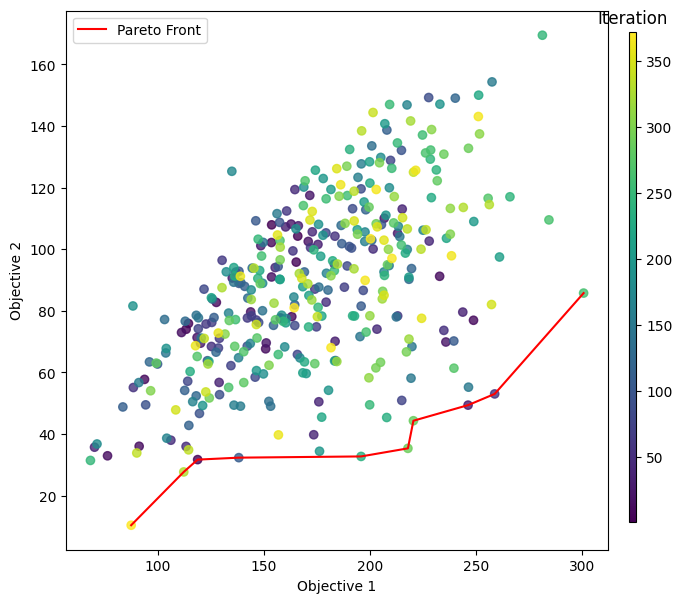} & \includegraphics[width=0.45\linewidth]{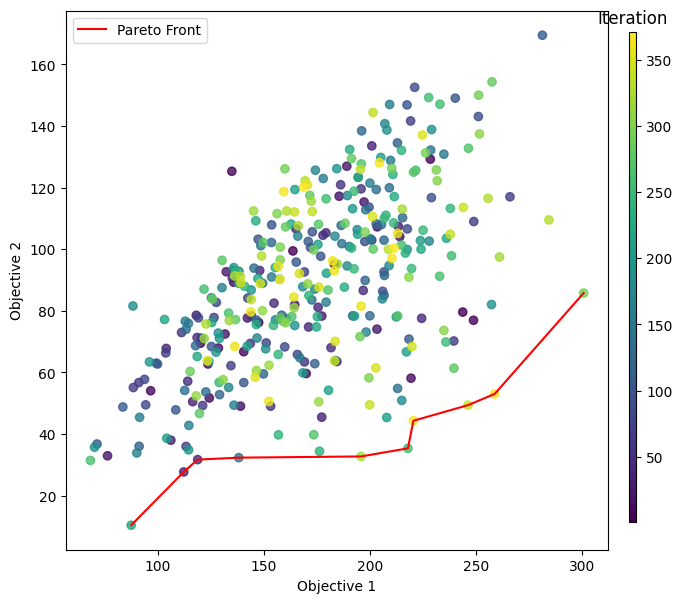} \\
    UDS & SPM \\
    Data Points Needed: 92.54\% & Data Points Needed: 92.29\%
  \end{tabular}
  \caption{Comprehensive Comparison Among the Models with respect to Data Usage (D)}
  \label{tab:fig12}
\end{figure}

Based on Fig.~\ref{tab:fig12}, it can be seen that compared to the max-max scenario, BMSDM requires more data to display a real Pareto front, but it still requires less data than the other three space-filling techniques. To achieve a true Pareto front, all of the space-filling DoE requires more than 90\% of the data. SPM and UDS exhibit comparable results, with SPM having slightly fewer data. Once more, LHS does poorly in this situation and requires more than 98\% of the data. In the following, we investigate the number of iterations required to reach satisfied performance metrics by closely examining the algorithm and splitting the iteration into four parts: the first iteration, 25\% iteration, 50\% iteration, and 100\% iteration.
\begin{figure}[!ht]
  \centering
  \begin{tabular}{cc}
    \includegraphics[width=0.45\linewidth]{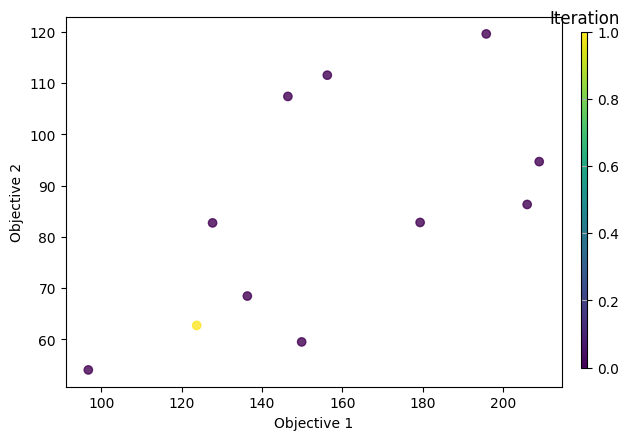} & \includegraphics[width=0.45\linewidth]{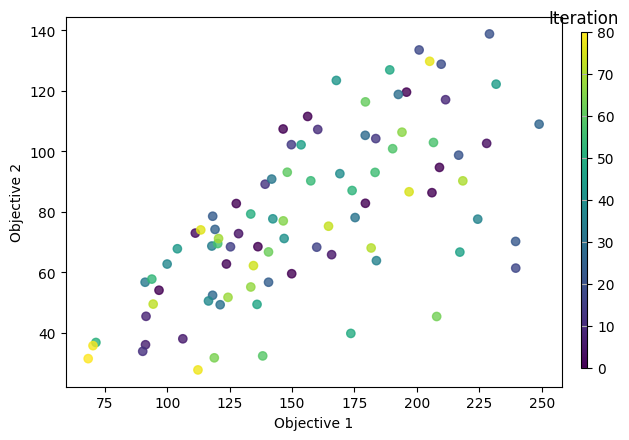} \\
    1$^\text{st}$ Iteration: & 25\% Iteration: \\
    GD: 42.98 \quad IGD: 41.81 & GD: 24.24 \quad IGD: 15.21 \\
    HV: 54.55\% \quad PHV: 57.72\% & HV: 77.33\% \quad PHV: 81.82\% \\
    \includegraphics[width=0.45\linewidth]{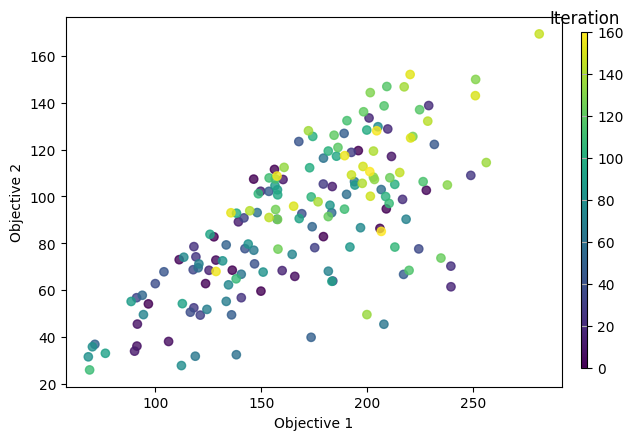} & \includegraphics[width=0.45\linewidth]{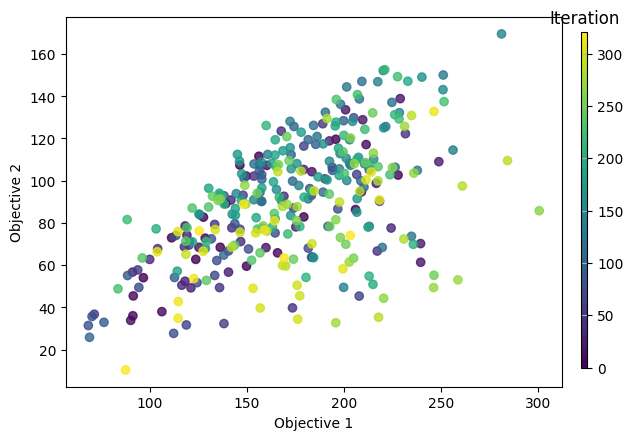} \\
    50\% Iteration: & 100\% Iteration: \\
    GD: 38.58 \quad IGD: 26.95 & GD: 0.004 \quad IGD: 0.004 \\
    HV: 78.58\% \quad PHV: 83.14\% & HV: 94.52\% \quad PHV: 100\%
  \end{tabular}
  \caption{Performance Analysis of BMSDM After Different Numbers of Iterations}
  \label{tab:fig13}
\end{figure}

Based on Fig.~\ref{tab:fig13}, it can be said that more than 80\% of the PHV value can be achieved within 25\% iteration. All of the true Pareto points are obtained after 100\% iteration. Therefore, we can limit the model to 25\% iterations which is more economical. Similar to the max-max scenario, beyond 25\% of iterations, further computational costs did not yield significant improvements in performance. Therefore, we restrict the data usage to 100 points (D = 25\%) and compare the proposed model with the space-filling methods in terms of the other four metrics. The algorithm achieves its goal by starting with an initial sample outside the Pareto front, selecting 10 points randomly to create the surrogate model. Our framework will choose the other 90 sequential candidate points autonomously.

\begin{figure}[!ht]
  \centering
  \begin{tabular}{cc}
    \includegraphics[width=0.45\linewidth]{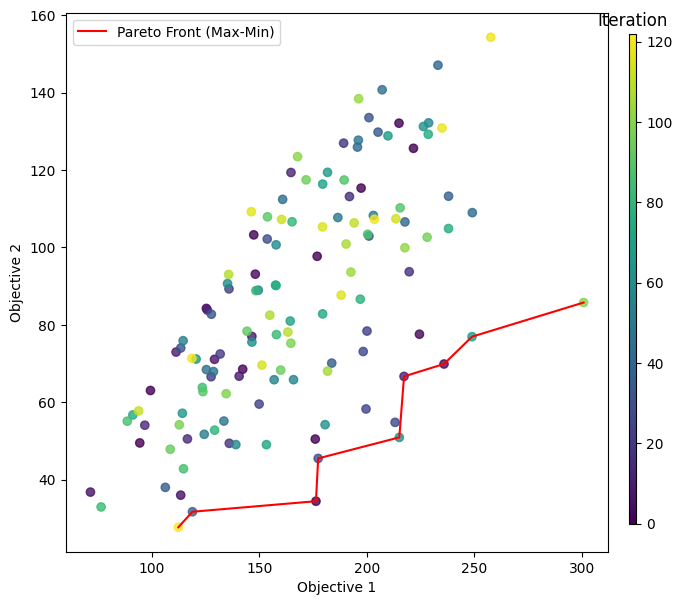} & \includegraphics[width=0.45\linewidth]{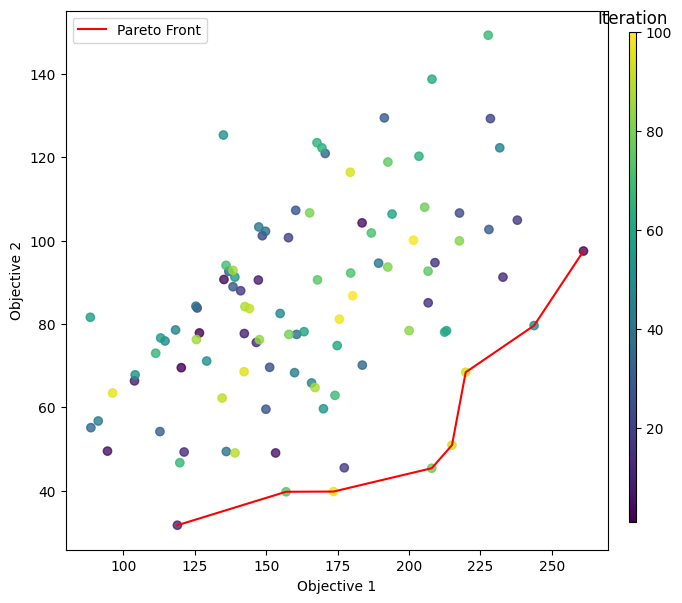} \\
    BMSDM & LHS \\
    GD: 17.25 \quad IGD: 13.59 & GD: 22.49 \quad IGD: 17.11 \\
    HV: 0.86 \quad PHV: 91.15\% & HV: 0.77 \quad PHV: 81.81\% \\
    \includegraphics[width=0.45\linewidth]{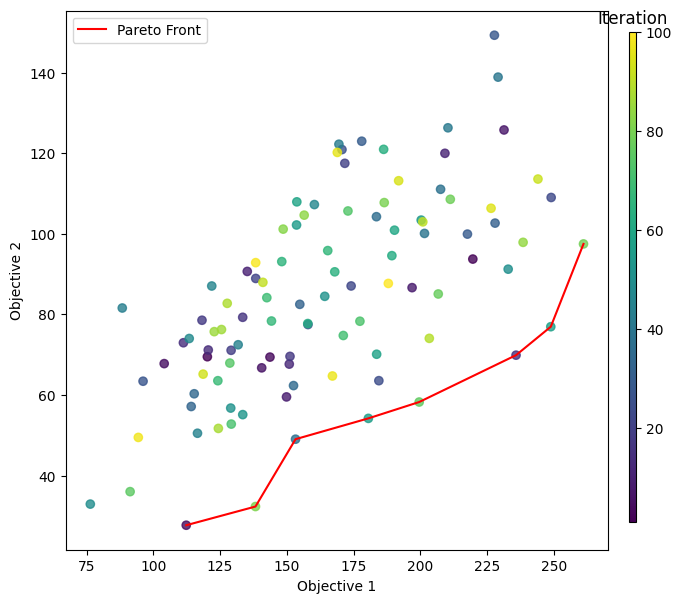} & \includegraphics[width=0.45\linewidth]{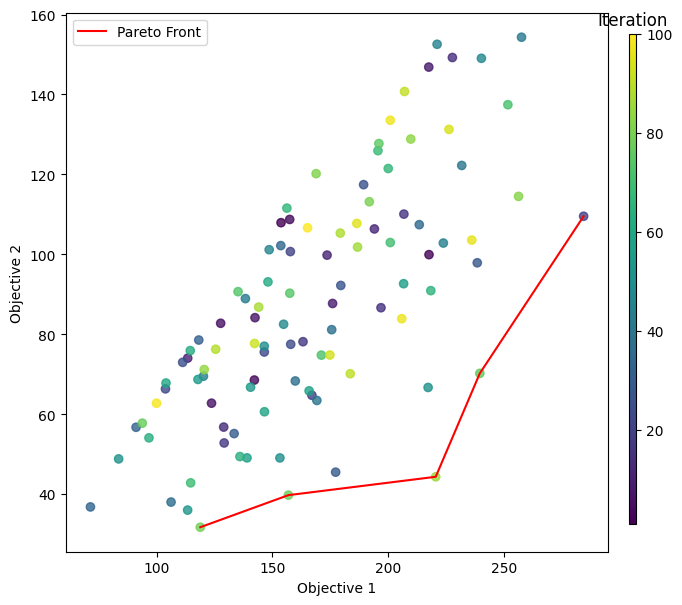} \\
    UDS & SPM \\
    GD: 23.43 \quad IGD: 20.05 & GD: 18.51 \quad IGD: 14.16 \\
    HV: 0.77 \quad PHV: 81.71\% & HV: 0.8 \quad PHV: 84.6\%
  \end{tabular}
  \caption{Comprehensive Comparison Among the Models Using Performance Metrics}
  \label{tab:fig14}
\end{figure}

{Fig.~\ref{tab:fig14} clearly demonstrates that BMSDM outperforms the others in every assessed category. UDS and LHS show relatively similar results across all metrics. SPM performs better than these two methods. Only BMSDM achieves more than 90\% PHV in this scenario. BMSDM also illustrates lower GD and IGD values, which indicates that this framework achieves a more accurate and well-distributed approximation of the true Pareto-optimal front compared to the other methods. To obtain a more reliable evaluation of the GD, IGD, HV, and PHV values and to demonstrate the models' stability, twenty-five runs are conducted.
\begin{figure}[!ht]
  \centering
  \begin{tabular}{cc}
    \includegraphics[width=0.45\linewidth]{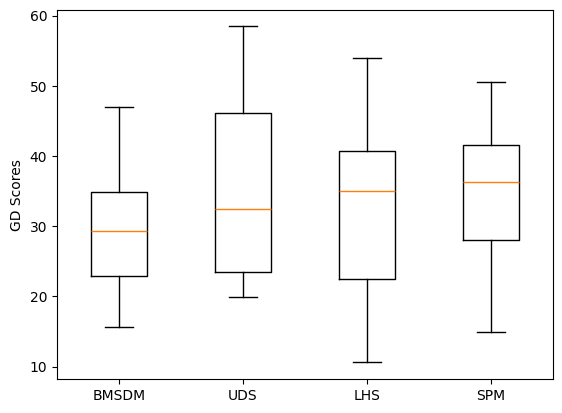} & \includegraphics[width=0.45\linewidth]{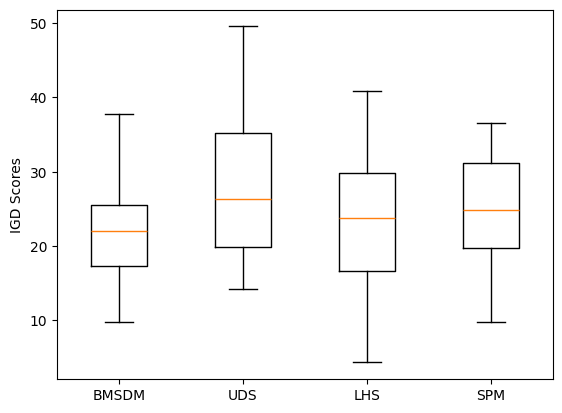} \\
    GD & IGD \\
    \includegraphics[width=0.45\linewidth]{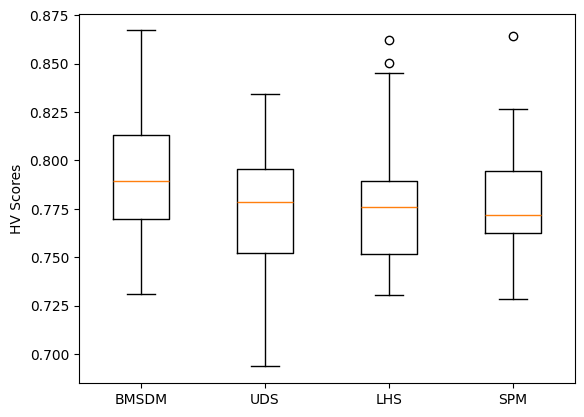} & \includegraphics[width=0.45\linewidth]{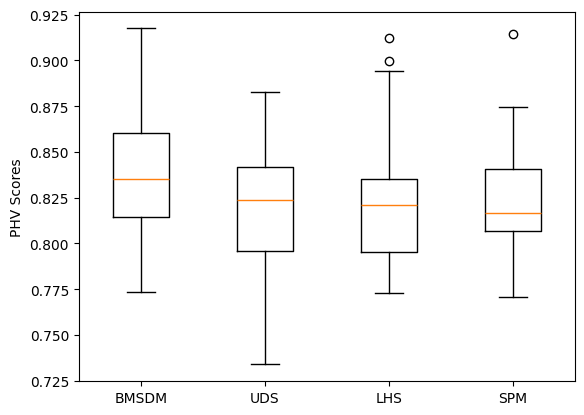} \\
    HV & PHV
  \end{tabular}
  \caption{Stability of the Models Based on Different Performance Metrics}
  \label{tab:fig15}
\end{figure}

The box plots in Fig.~\ref{tab:fig15} clearly show that BMSDM outperforms all competing methods across every performance metric. Even in max-min scenarios, where achieving optimum results in competitive objectives requires more data, BMSDM remains the leader. BMSDM achieves the lowest GD and IGD, which indicates that its solutions are closest to and well-distributed along the Pareto front. It also achieves the highest HV and PHV, meaning it captures a larger portion of the Pareto front and provides better trade-off solutions. The performance of all space-filling methods is nearly identical; UDS performs the worst, showing the highest IGD with large variability, indicating poor convergence and inconsistent results. LHS and SPM perform moderately, with LHS slightly outperforming SPM in capturing diversity and trade-offs. LHS has slightly lower variability in IGD distributions, while SPM shows a higher median IGD value. LHS has some outliers, indicating that it might find better solutions in some cases. BMSDM has a compact distribution in all the performance metrics, indicating it is more consistent across different runs. Overall, BMSDM emerges as the most reliable and effective method for the max-min scenario too, ensuring both high-quality solutions and better front coverage.

\subsection{Comparison with Other State-of-the-Art Approaches} 

To demonstrate the superiority of our proposed approach, we compare it with two advanced MOO methods: Non-dominated Sorting Genetic Algorithm II (NSGA-II) \citep{deb2002fast}, a metaheuristic approach, and Knowledge-Guided Bayesian Dynamic Multi-Objective Evolutionary Algorithm (KGB-DMOEA) \citep{ye2022knowledge}, a ML-based method. NSGA-II, one of the most widely used evolutionary algorithms, improves upon its predecessor, NSGA, by integrating elitism, which helps retain previously identified Pareto-optimal solutions. It replaces the traditional sharing mechanism with crowding distance, offering a more efficient, parameter-free strategy for maintaining solution diversity \citep{abdel2018metaheuristic}. However, NSGA-II has notable limitations, particularly in complex, multimodal landscapes where it may prematurely converge to a suboptimal Pareto front due to inadequate exploration. Being a population-based algorithm, it requires numerous function evaluations across multiple generations, making it computationally expensive. Its effectiveness also heavily depends on parameter tuning, including population size, crossover, and mutation rates. Additionally, NSGA-II does not utilize past evaluations, often leading to redundant searches even when similar solutions have already been explored \citep{kareem2022metaheuristic, abdel2018metaheuristic}.

KGB-DMOEA, on the other hand, draws inspiration from human learning behavior, where knowledge is accumulated, reconstructed, analyzed, and applied to new problems. It classifies past search experiences into valuable and non-valuable knowledge, allowing it to adapt more efficiently to dynamic environments. Using a knowledge reconstruction-examination mechanism alongside a naïve Bayesian classifier, it leverages historical search data to improve the prediction of high-quality initial populations \citep{ye2022knowledge}. By exploiting past information, KGB-DMOEA can estimate optimal solutions more effectively. However, unlike BMSDM, it does not follow a strictly sequential approach. Instead, it employs an evolutionary or hybrid process that generates solutions in batches, enhancing diversity but at the cost of higher function evaluation overhead compared to sequential BO. The following subsections analyze the results of NSGA-II and KGB-DMOEA applied to our MOO problem, evaluating their performance against key metrics and comparing them with our proposed framework under both optimization scenarios.

\subsubsection{ Max-Max Scenario}

\begin{figure}[!ht]
  \centering
  \begin{tabular}{cc}
    \includegraphics[width=0.45\linewidth]{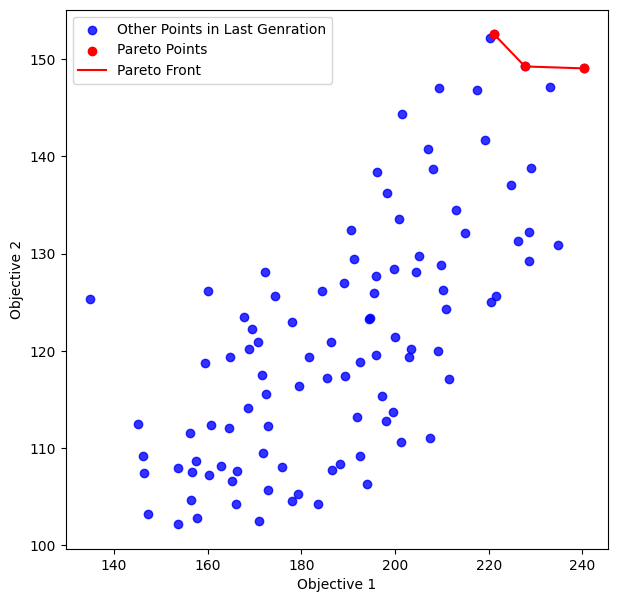} & \includegraphics[width=0.45\linewidth]{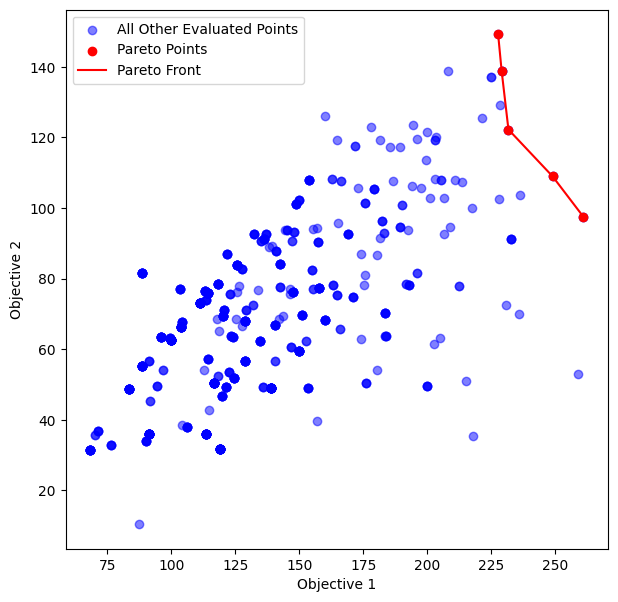} \\
        NSGA-II & KGB-DMOEA \\
  \end{tabular}
  \caption{Pareto Front of NSGA-II and KGB-DMOEA (Max-Max Scenario)}
  \label{tab:fig16}
\end{figure}

The Pareto front obtained using NSGA-II and KGB-DMOEA are illustrated in Fig.~\ref{tab:fig16}. Both of the algorithms miss the true Pareto points and provide almost similar results. Compared with BMSDM (Fig.~\ref{tab:fig8} and Fig.~\ref{tab:fig9}), the quality of the Pareto front is pretty inferior.

%\begin{table}[!ht]
\begin{table}[!ht]
    \centering
    \caption{Performance Evaluation of NSGA-II, KGB-DMOEA, and BSDM for Max-Max Scenario (Bold Indicates the Best Performance)}
    \label{tab:max_max}
    \renewcommand{\arraystretch}{1.3}
    \begin{tabular}{|l|c|c|c|c|c|}
        \hline
        & \textbf{GD} & \textbf{IGD} & \textbf{HV} & \textbf{PHV} & \textbf{Function Evaluation/Data Usage} \\
        \hline
        \textbf{NSGA II} & 55.64 & 55.2 & 0.72 & 74.04\% & 1000 \\
        \hline
        \textbf{KGB-DMOEA} & 48.61 & 46.71 & 0.74 & 76.01\% & 730 (172 Unique Evaluations) \\
        \hline
        \textbf{BMSDM} & \textbf{0.0047} & \textbf{0.0047} & \textbf{0.93} & \textbf{96.48\%} & \textbf{56} \\
        \hline
    \end{tabular}
\end{table}

Table~\ref{tab:max_max} summarizes the performance of all three algorithms. BMSDM significantly outperforms NSGA-II and KGB-DMOEA across all metrics, demonstrating superior convergence and diversity. With less than 10\% of NSGA-II’s function evaluations and 30\% of KGB-DMOEA’s unique solutions, BMSDM achieves over 90\% of the total hypervolume. Unlike NSGA-II and KGB-DMOEA, which rely on random mutation and crossover, BMSDM employs probabilistic models for solution refinement, ensuring rapid discovery of high-quality, diverse solutions (high HV, PHV) without excessive function evaluations. The drastically lower GD and IGD values further confirm that BMSDM’s solutions are closer to the true Pareto front.

NSGA-II, with a population size of 100, 20\% crossover/mutation rates, and a 5\% local search rate, requires 1000 function evaluations over 10 generations to approximate the Pareto front \citep{blank2020pymoo}. These parameters are carefully tuned to keep predicted solutions within dataset boundaries \citep{wang2019parameterization}. While increased function evaluations could improve accuracy, such an approach is impractical due to the resource-intensive nature of experimental validation. In contrast, KGB-DMOEA achieves comparable results in just four generations, reducing total function evaluations by nearly half, and even more in terms of unique evaluations. This efficiency stems from its Bayesian learning mechanism, which retains a historical record of optimal solutions, eliminating redundant evaluations and accelerating convergence. Moreover, KGB-DMOEA is less dependent on parameter tuning than NSGA-II, which requires careful calibration of population size, crossover, and mutation rates.

Despite its advantages over NSGA-II, KGB-DMOEA remains computationally expensive due to its reliance on evolutionary processes, which require evaluating a larger population. BMSDM, by contrast, is more sample-efficient, selecting optimal samples based on uncertainty quantification through Bayesian optimization. By balancing exploration and exploitation via acquisition functions, BMSDM achieves faster convergence with significantly fewer function evaluations.

\subsubsection{ Max-Min Scenario}

Fig.\ref{tab:fig17} presents the Pareto front of NSGA-II and KGB-DMOEA in the max-min scenario. Both algorithms exhibit improved Pareto fronts compared to the max-max scenario. However, NSGA-II fails to capture 6 true Pareto points, while KGB-DMOEA misses 2, indicating a performance gap. When compared to BMSDM (Fig.\ref{tab:fig12} and Fig.~\ref{tab:fig13}), NSGA-II demonstrates the weakest Pareto front, while KGB-DMOEA performs better but still does not match the quality and accuracy of BMSDM.

\begin{figure}[!ht]
  \centering
  \begin{tabular}{cc}
    \includegraphics[width=0.45\linewidth]{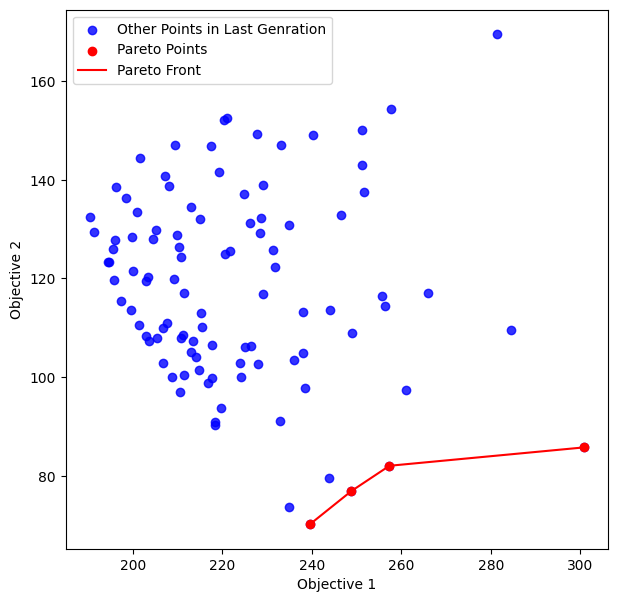} & \includegraphics[width=0.45\linewidth]{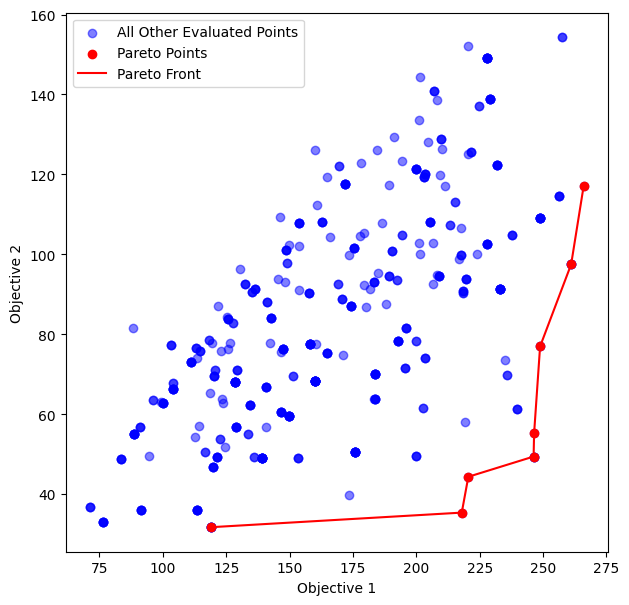} \\
        NSGA-II & KGB-DMOEA \\
  \end{tabular}
  \caption{Pareto Front of NSGA-II and KGB-DMOEA (Max-Min Scenario)}
  \label{tab:fig17}
\end{figure}

Table~\ref{tab:max_min} compares the three strategies based on performance metrics. While BMSDM’s performance slightly declines, it still outperforms both NSGA-II and KGB-DMOEA across all measures, demonstrating its ability to capture more true Pareto points (low GD and IGD values) with better convergence and fewer evaluations. BMSDM achieves 21\% more total hypervolume while using around 1\% of NSGA-II’s function evaluations. Leveraging sequential point selection, BMSDM attains approximately 6\% higher hypervolume than KGB-DMOEA while requiring fewer data points, reinforcing its efficiency.

\begin{table}[!ht]
    \centering
    \caption{Performance Evaluation of NSGA-II, KGB-DMOEA, and BSDM for Max-Min Scenario (Bold Indicates the Best Performance)}
    \label{tab:max_min}
    \renewcommand{\arraystretch}{1.3}
    \begin{tabular}{|c|c|c|c|c|c|}
        \hline
        & \textbf{GD} & \textbf{IGD} & \textbf{HV} & \textbf{PHV} & \textbf{Function Evaluation/Data Usage} \\
        \hline
        \textbf{NSGA II} & 22.33 & 19.21 & 0.67 & 70.64\% & 10000 \\
        \hline
        \textbf{KGB-DMOEA} & 23.97 & 15 & 0.81 & 85.85\% & 520 (172 Unique Evaluations) \\
        \hline
        \textbf{BMSDM} & \textbf{17.25} & \textbf{13.59} & \textbf{0.86} & \textbf{91.15}\% & \textbf{132} \\
        \hline
    \end{tabular}
\end{table}

NSGA-II, with the same 100 starting population, an 85\% crossover and 10\% mutation rate, and a 5\% local search rate, requires over 10000 function evaluations across 1000 generations to approximate the Pareto-optimal front \citep{ferreira2023nsga}. This enormous amount of function evaluation makes it an extremely inefficient algorithm in building trials of high cost. In contrast, KGB-DMOEA provides superior performance in just three generations while reducing the total function evaluations substantially. While KGB-DMOEA improves upon typical evolutionary multi-objective techniques by integrating adaptive sampling, it still operates in a population-based manner rather than a purely sequential fashion, making it less efficient than BMSDM.

\section{Conclusion}
\label{Conclusion}
In this study, we present BMSDM, an intelligent MOBO framework that marks a significant leap forward toward efficient experimental design for smart manufacturing and materials discovery. By leveraging a surrogate GP-based model for strategic data point selection across various design or production parameters, BMSDM offers a substantial enhancement over the traditional DoE methods such as LHS, UDS, and SPM. Our comparative analysis, employing metrics like GD, IGD, HV, PHV, and D, underlines the superior efficacy of BMSDM in navigating complex decision spaces. Utilizing a comprehensive manufacturing dataset, the framework excels in adaptability and precision, thriving particularly in scenarios that demand the maximization minimization, or trade-offs between objective functions. The significance of BMSDM transcends its immediate operational benefits, signaling a shift towards a more efficient, data-driven approach to manufacturing and material discovery. By minimizing resource and data requirements, it enables more cost-effective and swift advancements in these fields, which is vital for economic viability and expediting the innovation cycle.

BO is highly effective for low-dimensional black-box optimization but faces challenges in high-dimensional spaces due to exponential search space growth, complex model training, and rising computational costs. The cubic complexity of GP regression limits scalability, and while scalable GPs exist, they primarily address large datasets rather than high-dimensional optimization. Moreover, as dimensionality increases, optimizing the acquisition function and hyperparameter tuning becomes increasingly computationally demanding, with qEHVI’s efficiency constrained by its partitioning approach. Improving partitioning strategies could enhance scalability. Additionally, batch BO presents execution challenges, such as maintaining stable temperature settings across samples while ensuring sufficient variability in parameters like length.

Looking ahead, our research will explore these limitations and find a solution to create a more robust scalable BO algorithm. The integration of traditional DoE methods with BO to optimize initial point selection can be a good future research direction. This integration aims to further enhance operational efficiency and tackle the challenge of local optima by exploring novel acquisition functions. Unlike fixed schedules or static setups, this intelligent algorithm can work autonomously by identifying maintenance needs, adapting quality control protocols to material changes, and dynamically updating production schedules in real time. These functions are anticipated to refine the model's ability to discover a wider range of optimal solutions, thereby enhancing the strategic capability of this promising optimization framework in transforming industry practices. To augment the practical usefulness of BO in manufacturing, subsequent efforts should concentrate on enhancing high-dimensional scalability, refining acquisition function assessment, and devising more adaptable batch selection procedures. Leveraging contemporary computational paradigms and technology could further alleviate these restrictions, making BO more realistic for large-scale industrial applications.
\FloatBarrier  % Ensure all floats are processed before the references
\vspace{20pt} % This adds a section-like gap
\noindent \textbf{Data Availability} The datasets generated and analysed during the current study are available from the corresponding author upon reasonable request.

\section*{Statement \& Declarations}
\begin{itemize}
\item \textbf{Competing interests} The authors have no competing interests to declare that are relevant to the content of this article.
\item \textbf{Financial support} The authors did not receive financial support from any organization for the submitted work.
\item \textbf{Consent to participate} All authors have consent to participate.
\item \textbf{Consent for publication} All authors have consent for publication.
\item \textbf{Ethical Approval} Not Applicable.
\end{itemize}

%\bibliography{sn-bibliography}
\bibliography{references}

\end{document}